\documentclass[10pt,twocolumn,letterpaper]{article}

\usepackage{iccv}

\usepackage[accsupp]{axessibility}  
\usepackage{times}
\usepackage{epsfig}
\usepackage{graphicx}
\usepackage{amsmath}
\usepackage{amssymb}
\usepackage{booktabs}
\usepackage{xr-hyper}
\usepackage{caption}
\usepackage{subcaption}
\usepackage{multirow}
\usepackage{array}
\usepackage{arydshln}

\usepackage{siunitx}
\usepackage[breaklinks=true,bookmarks=false]{hyperref}

\iccvfinalcopy 

\usepackage[capitalize]{cleveref}
\crefname{section}{Sec.}{Secs.}
\Crefname{section}{Section}{Sections}
\Crefname{table}{Table}{Tables}
\crefname{table}{Tab.}{Tabs.}

\def\iccvPaperID{11568}

\begin{document}

\title{VQA Therapy: Exploring Answer Differences by Visually Grounding Answers}

\author{\hspace{-0.1in} Chongyan Chen$^1$, Samreen Anjum$^2$, Danna Gurari$^1$$^,$$^2$ \\
\noindent
{\hspace{-0.2in} \small $~^1$ University of Texas at Austin}
{\small $~^2$ University of Colorado Boulder}
}
\maketitle
\begin{abstract}
Visual question answering is a task of predicting the answer to a question about an image.  Given that different people can provide different answers to a visual question, we aim to better understand why with answer groundings.  We introduce the first dataset that visually grounds each unique answer to each visual question, which we call VQA-AnswerTherapy.  We then propose two novel problems of predicting whether a visual question has a single answer grounding and localizing all answer groundings. We benchmark modern algorithms for these novel problems to show where they succeed and struggle.  The dataset and evaluation server can be found publicly at 
https://vizwiz.org/tasks-and-datasets/vqa-answer-therapy/. 
 \end{abstract}
\section{Introduction}
Visual question answering (VQA) is the task of predicting the answer to a question about an image. A fundamental challenge is how to account for when a visual question has multiple natural language answers, a scenario shown to be common~\cite{gurari2017crowdverge}.  Prior work~\cite{bhattacharya2019does} revealed reasons for these differences, including due to subjective or ambiguous visual questions. However, it remains unclear to what extent answer differences arise because \emph{different visual content} in an image is described versus because the \emph{same visual content} is described differently (e.g., using different language).  

\begin{figure}[t!]
     \hspace{-2.2em}
     \includegraphics[width=0.55\textwidth]{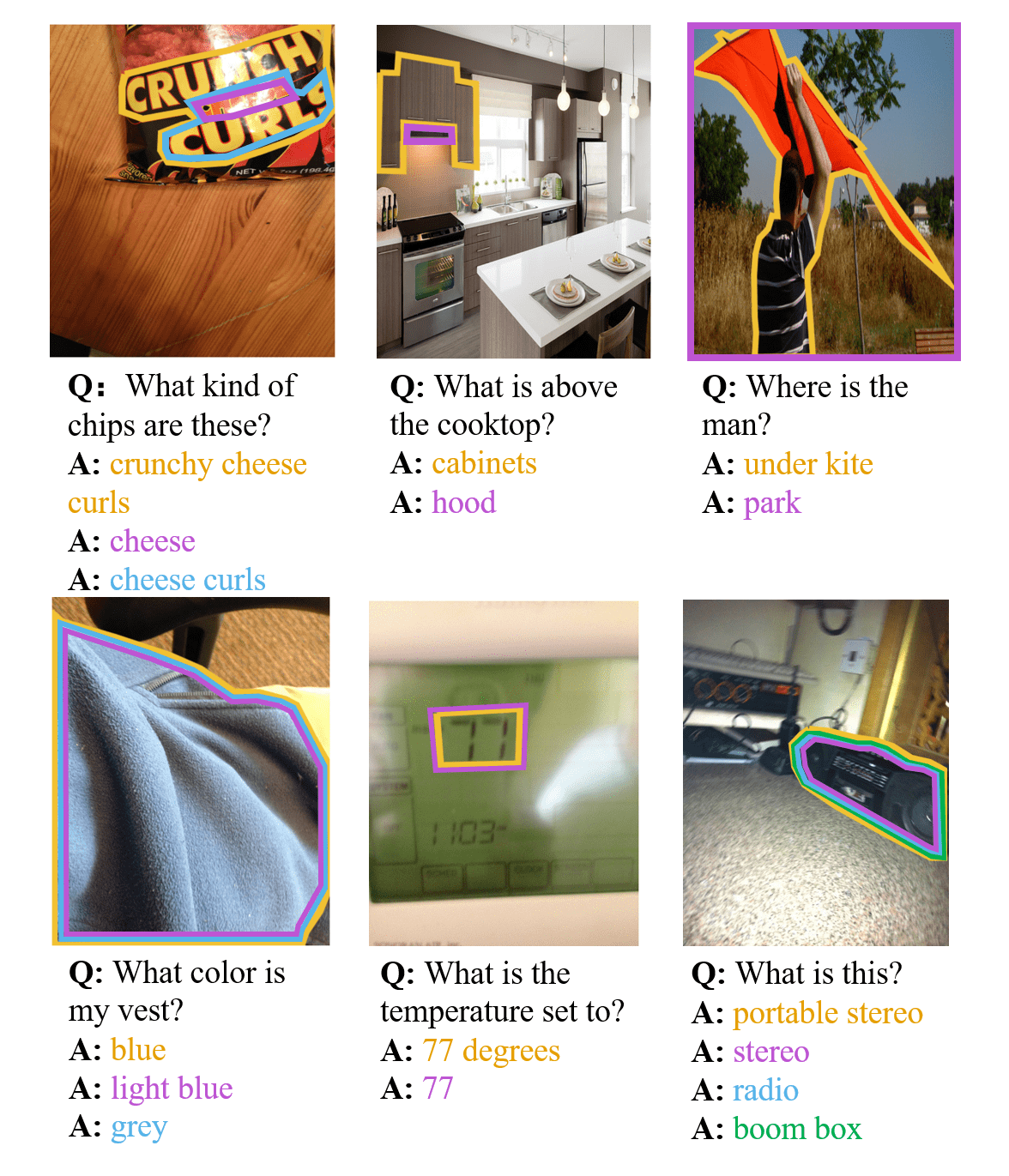}
     \vspace{-2em}
        \caption{Examples from our VQA-AnswerTherapy dataset showing that visual questions with different natural language answers can have multiple answer groundings (first row) or all share the same answer grounding (second row).}
    \label{fig:main}
\end{figure}

Our work is designed to disentangle the vision problem from other possible reasons that could lead to answer differences.  To do so, we introduce the first dataset where all valid answers to each visual question are grounded, meaning we segment for each answer the visual evidence in the image needed to arrive at that answer.  This new dataset, which we call VQA-AnswerTherapy, consists of 5,825 visual questions from the popular VQAv2 \cite{balanced_vqa_v2} and VizWiz \cite{gurari2018vizwiz} datasets. We find that 16\% of the visual questions have multiple answer groundings, and provide fine-grained analysis to better elucidate when and why this arises. 

We also introduce two novel algorithmic challenges, which are exemplified in Figure \ref{fig:main}.  First, we introduce the \textbf{Single Answer Grounding Challenge}, which entails predicting for a visual question whether all valid answers will describe the same grounding or not. Next, the \textbf{Grounding Answer(s) Challenge} entails localizing the answer groundings for all valid answers to a visual question. We benchmarked models for these novel tasks to demonstrate the baseline performance for modern architectures and to highlight where they are succeeding and struggling. 


We offer this work as a valuable foundation for improving our understanding and handling of annotator differences.  Success on our \textbf{Single Answer Grounding Challenge} will enable solutions to notify users when there is uncertainty which visual evidence to consider, enabling users to clarify a visually ambiguous visual question.  This can immediately benefit visually impaired individuals since a portion of our dataset comes from this population (i.e., VizWiz \cite{gurari2018vizwiz}). Success on the \textbf{Answer(s) Grounding Challenge} can enable users of VQA solutions to better understand the varied reasoning processes that can lead to different natural language answers, while also contributing to enhanced model explainability and trustworthiness.  More generally, this work can inform how to account for \textbf{annotator differences} for other related tasks such as image captioning, visual dialog, and open-domain VQA (e.g., VQAs found on Yahoo!Answers and Stack Exchange).  This work also contributes to ethical AI by enabling revisiting how VQA models are developed and evaluated to consider the diversity of plausible answer groundings rather than a single (typically majority) one.  To facilitate future extensions, we publicly-share our dataset and a public evaluation server with leaderboard for our dataset challenges at the following link: 
 https://vizwiz.org/tasks-and-datasets/vqa-answer-therapy/.



\section{Related Work}

\paragraph{Answer Differences in VQA Datasets.}
While many datasets have been created to support the development of VQA algorithms~\cite{singh2019towards,balanced_vqa_v2,gurari2018vizwiz}, a long-standing challenge has been how to account for the common situation that, for many visual questions, different answers are observed from different people~\cite{gurari2017crowdverge}.  Prior work has offered initial steps.  For example, prior work characterized when~\cite{gurari2017crowdverge}, to what extent~\cite{yang2018visual}, and why answers differ in mainstream VQA datasets (e.g., for visual questions that are difficult, ambiguous, or subjective as well as answers that are synonymous)~\cite{bhattacharya2019does}.  Other work introduced ways to evaluate VQA models that acknowledge there can be multiple valid answers, whether provided explicitly from different people~\cite{antol2015vqa,luo2021just} or augmented automatically from NLP tools to capture plausible, semantically related answers~\cite{luo2021just}. Another work focused on rewriting visual questions to remove ambiguity regarding what are valid answers~\cite{stengel2022did}. Complementing prior work, we explore answer differences in the VQA task from the perspective of grounding, specifically exploring whether different answers arise because \emph{different visual content} in an image is being described.
  
\vspace{-1em}\paragraph{Answer Grounding Datasets. }
Numerous datasets have been proposed to support developing models that locate the visual evidence humans rely on to answer visual questions.  This has been motivated by observations that answer groundings can serve as a valuable foundation for debugging VQA models, providing explanations for VQA model predictions, protecting user privacy by enabling obfuscation of irrelevant content in images, and facilitating search behaviors by highlighting relevant visual content in images.  A commonality of prior work~\cite{gan2017vqs,nagaraj-rao-etal-2021-first,huk2018multimodal,das2017human,chen2020air,Zhu_2016_CVPR, krishna2017visual,hudson2019gqa, huk2018multimodal,nagaraj-rao-etal-2021-first,chen2022grounding,Chen_2022_CVPR,wu2020phrasecut} is that only one answer grounding for one selected answer is provided for each visual question. Our work, in contrast, acknowledges that a visual question can have multiple valid answers and so multiple valid answer groundings. We introduce the first dataset where all valid answers to each visual question are grounded. This new dataset, which we call VQA-AnswerTherapy, enables us to introduce two novel tasks of predicting for a given visual question whether all answers will be based on the same visual evidence and predicting for a visual question the groundings for all valid answers.

\vspace{-1em}\paragraph{Automated VQA Methods.}
Modern automated VQA models typically only return a single answer; e.g., the predicted answer with the highest probability from a softmax output layer of a neural network.  Yet, people often ask visual questions that lead to multiple valid answers~\cite{gurari2017crowdverge}.  To account for this practical reality, we propose novel tasks and introduce the first models for sharing richer information with end users by (1) indicating when there are multiple plausible answer groundings to a visual question and (2) locating those grounded regions in images.

\section{VQA-AnswerTherapy Dataset}

\subsection{Dataset Creation}
\label{section:datasetcreation}

\paragraph{VQA Source.}
Our work builds upon two popular VQA datasets that reflect two distinct scenarios: VizWiz-VQA~\cite{gurari2018vizwiz} and VQAv2~\cite{balanced_vqa_v2}.  The images and questions of the VizWiz-VQA dataset come from visually impaired people who shared them in authentic use cases where they were trying to obtain assistance in understanding their visual surroundings.  In contrast, the images and questions of the VQAv2 dataset come from different sources: while the images come from the MS COCO dataset \cite{chen2015microsoft} (and so were collected from the Internet), the questions were generated by crowd workers.  Despite these differences, these datasets have in common that they both include for each image-question pair 10 crowdsourced answers, each of which was curated based on the same crowdsourcing interface. 

\vspace{-1em}\paragraph{VQA Filtering.}  
Our goal is to unambiguously ground each answer for \emph{visual questions that have more than one valid answer}.  To focus on these visual questions of interest, we applied filters to the original VQA sources, which consist of 32,842 image-question pairs for VizWiz-VQA and 443,757 for VQAv2 training dataset.  First, we removed answers indicating a visual question is unanswerable (i.e.,  ``unsuitable" or ``unanswerable").  Then, we only focused on the remaining visual questions that have two or more valid natural language answers, where we define valid answers as those for which at least two out of the ten crowdworkers gave the exact same answer (i.e., using string matching).  \footnote{We follow the status quo established by prior work \cite{chen2022grounding,gurari2017crowdverge} to obtain valid answers by using exact string matching (ESM) to provide an upper bound for expected differences). Around 36\% of visual questions in VizWiz and VQAv2 datasets have more than one \emph{valid} answer.}  Similar to prior work~\cite{chen2022grounding}, we also filter visual questions that embed multiple sub-questions. An example is ``How big is my TV and what is on the screen, and what is the model number, and what brand is it?"  Following \cite{chen2022grounding}, we removed visual questions with more than five words and the word ``and", trimmed visual questions containing a repeated question down to a single question ( e.g., from ``what is this? what is this?" to ``what is this?"), and filtered visual questions flagged as ``containing more than one question" in metadata provided by \cite{chen2022grounding}.

We then selected 27,741 visual questions with 60,526 unique answers as candidates for our dataset.  Included are all visual questions from VizWiz-VQA that met the aforementioned criteria (i.e., 9,528 visual questions with 20,930 unique answers) and a similar amount sampled from VQAv2's training set (i.e., 18,213 visual questions with 39,596 unique answers).  We included all visual questions used in \cite{bhattacharya2019does}, which indicates why answers to each visual question differ, to support downstream analysis.  

\vspace{-1em}\paragraph{Answer Grounding Task Design.} 
We designed a user interface to ground the different answers for each visual question. It presents the image-question pair alongside one of its associated answers at a time.  

For each answer, two questions were asked to ensure the answer could be unambiguously grounded to one region.  First, a worker had to indicate if a given answer is correct or not.  If correct, then the worker had to specify how many polygons must be drawn to ground the answer from the following options: zero, one, or more than one.  To simplify the task, we only instructed the worker to ground the answer when exactly one polygon was needed to ground answer.  We leave future work to explore when there are multiple polygons (e.g., ``How much money is there?" for an image showing multiple coins).

To ground an answer, a worker was instructed to click a series of points on the image to create a connected polygon.  After one answer grounding was generated for a visual question, the annotator could then choose for a new answer to select a previously drawn polygon or draw a new polygon. Instructions were provided for how to complete the task, including for many challenging annotation scenarios (e.g., objects with holes or complex boundaries).

\begin{figure}[b!]
     \centering
     \includegraphics[width=0.47\textwidth]{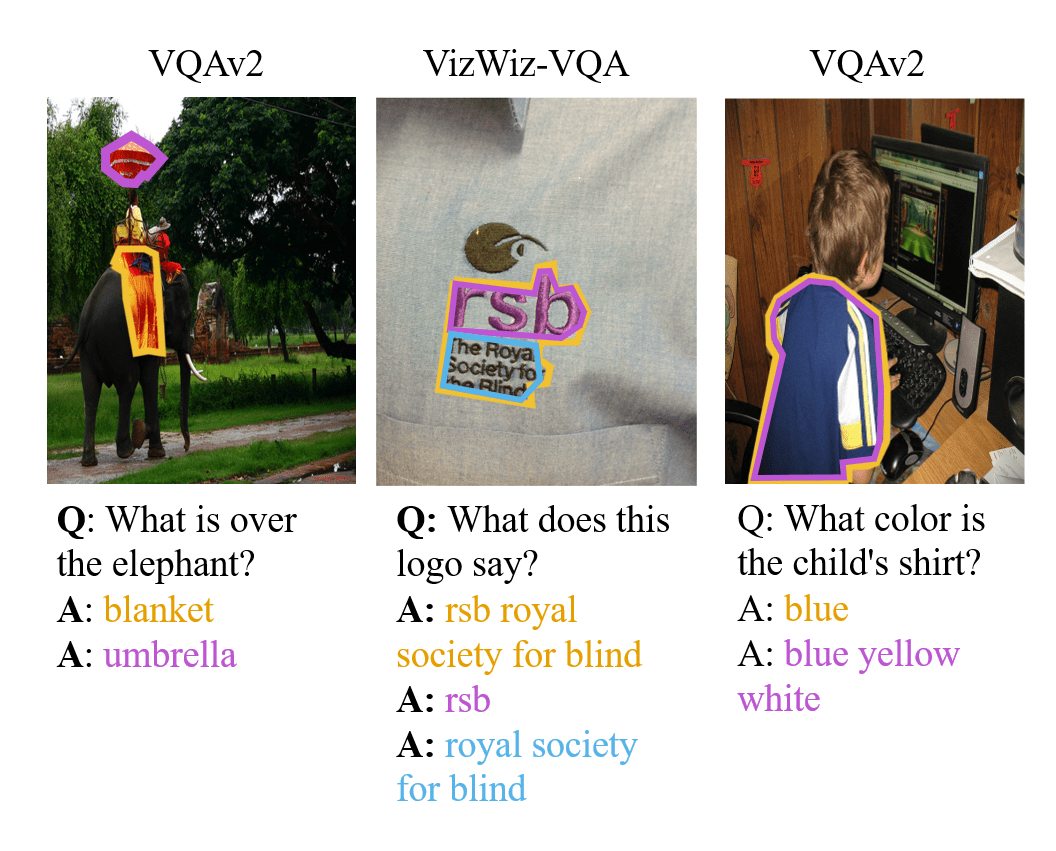}
     \vspace{-1.5em}
        \caption{High-quality annotations from our dataset.  These also illustrate a trend that visual questions related to \textit{text recognition} often have {multiple answer groundings} while \textit{recognizing color} often have a \emph{single grounding}. }
    \label{fig:highquality}
\end{figure}

\vspace{-1em}\paragraph{Answer Grounding Annotation Collection.}
We hired crowd workers from Amazon Mechanical Turk to generate answer groundings, given their on-demand availability.  Like prior work~\cite{chen2022grounding}, we only accepted workers from the United States who had completed at least 500 Human Intelligence Tasks (HITs) with over a 95\% acceptance rate. For each candidate worker, we provided a one-on-one zoom training on our task.  We then provided a qualifying annotation test to verify workers understood the instructions, and only accepted workers who passed this test. 

For annotation of our VQAs, we collected two answer groundings per image-question-answer triplet to enable examination of whether the annotations match and so are likely unambiguous, high-quality results. To support the ongoing collection of high-quality results, we also conducted both manual and automated quality control mechanisms.

\vspace{-1em}\paragraph{Ground Truth Generation. }
We analyzed the two sets of annotations collected for each of the 27,741 visual questions to establish ground truths.  We did this after removing answers that at least one person flagged as ``incorrect" and visual questions with answers referring to no polygon or multiple polygons.  This left 12,290 visual questions and 26,682 unique image-question-answer triplets. For each answer, we calculated the intersection-over-union (IoU) between the two answer groundings.  If the IoU was large (i.e., equal to or larger than 75\%), we used the larger of the two groundings as ground truth since often the smaller one is contained in the larger one.  Otherwise, we deemed that answer has an ambiguous grounding and so removed the answer from our dataset.  Examples of high-quality answer grounding results are shown in Figure \ref{fig:highquality}, where answers to a visual question can either have multiple groundings (e.g., ``What is over the elephant" and ``What does this logo say") or a single grounding (e.g., ``shirt's color"). 

\begin{table*}[t!]
\centering
\begin{tabular}{lllll}
\hline
     & & All                          & VQAv2                          & VizWiz-VQA                   \\ \hline
\multirow{5}{*}{\rotatebox[origin=c]{90}{\bf Multiple}} & Top-1 & What is this?                & What is the man wearing?       & What is this?                \\
& Top-2 & What is in this box?         & What is on the table?          & What is in this box?         \\
& Top-3 & What does this say?          & Where is the pizza?            & What does this say?          \\
& Top-4 & What is it?                  & What does the street sign say? & What is it?                  \\
& Top-5 & What kind of coffee is this? & What does the sign say?        & What kind of coffee is this? \\ \hline
\multirow{5}{*}{\rotatebox[origin=c]{90}{\bf Single}} &Top-1 & What is this?             & What color is the train? & What is this?             \\
& Top-2 & What color is this?       & What color is the cat?   & What color is this?       \\
& Top-3 & What is it?               & What is the man holding? & What is it?               \\
& Top-4 & What's this?              & What room is this?       & What color is this shirt? \\
& Top-5 & What color is this shirt? & What color is the bus?   & What color is this shirt? \\ \hline
\end{tabular}
\caption{The five most common questions that lead to \emph{multiple} answer groundings and a \emph{single} answer grounding for all visual questions as well as for VQAv2 and VizWiz-VQA independently. Of note, the overall frequency is dominated by VizWiz-VQA's frequency since the most common questions is far larger for this dataset than observed for VQAv2.}
\label{table:question-frequency}
\end{table*}

\subsection{Dataset Analysis}
\label{section:dataAnalysis}
We now analyze our final dataset, which includes 5,825 visual questions with 12,511 unique visual-question-answer-grounding sets. This includes 7,426 answer groundings for 3,442 visual questions from VizWiz-VQA dataset and 5,085 answer groundings for 2,383 visual questions from VQAv2 dataset.  This final dataset excludes all visual questions with less than two unique answers.  

\vspace{-1em}\paragraph{Prevalence of Single Versus Multiple Groundings.}
We first explore how often visual questions have different answers describing the same visual evidence (a single grounding) versus different visual evidence (multiple answer groundings). We flag a visual question as having different answers describing the \emph{same grounding} if an answer grounding pair has an IoU score larger than 0.9.  

We found 15.7\% (i.e., 916/5,825) of visual questions with answers leading to \emph{mulitiple answer groundings}. Yet, the status quo for VQA research neglects this reality that different answers can refer to different visual evidence~\cite{gan2017vqs,nagaraj-rao-etal-2021-first,huk2018multimodal,das2017human,chen2020air,Zhu_2016_CVPR, krishna2017visual,hudson2019gqa,gan2017vqs, huk2018multimodal,nagaraj-rao-etal-2021-first,chen2022grounding,Chen_2022_CVPR,wu2020phrasecut}. We suspect existing models would struggle with these 15.7\% questions, both for VQA and answer grounding, due to their visual ambiguity. 

We next identify the most common questions for visual questions that have \emph{multiple} as well as a \emph{single} answer grounding.  To do so, we tally how often each question leads to different answer groundings as well as to a single answer grounding respectively.  Results are shown in Table~\ref{table:question-frequency}.  We observe questions about \textit{recognizing objects} is common for both scenarios.  In contrast, questions about \textit{recognizing text} is more prevalent when there are \emph{multiple answer groundings} while questions about \textit{recognizing color} is more prevalent for visual questions with a \emph{single answer grounding}.  We also observe that questions related to a location often leads to multiple answer groundings, as shown in Table ~\ref{table:question-frequency} (Top-3 ``Where is the pizza") and exemplified in Figure \ref{fig:main} (``where is the man") and Figure \ref{fig:highquality} (``What is over the elephant"). 
These finding suggests that a valuable predictive cue for AI models to predict whether there is a single grounding or multiple groundings for all answers are identifying the vision skills needed to answer a visual question. 

When comparing the trend for visual questions to have multiple answer groundings across both data sources, we observe it is more prevalent for visual questions coming from VizWiz-VQA than VQAv2; i.e., it accounts for 22\% (i.e., 761/3,442) versus 7\% (i.e., 155/2,383) of visual questions respectively. Consequently, multiple answer groundings are more common for an authentic VQA use case than is captured in the most popular, yet contrived VQA dataset. 

\vspace{-1em}\paragraph{Reasons Visual Questions Have Multiple Answer Groundings.} 
We next analyze the 916 visual questions that have more than one answer grounding.  For each visual question, we flag which relationship types arise between every possible answer grounding pair from the following options: disjoint, equal, contained, and intersected.  We categorize an answer pair as \emph{disjoint} when IoU equals 0, \emph{equal} when the value is larger than 0.9, \emph{contained} when one region is part of the other region, such that the size of their intersection is equal to the minimum of their sizes and the size of their union is equal to the maximum of their sizes, and \emph{intersected} when 0.9$\geq$IoU $>$0 and they do not have a contained relationship. 

We first tally how many relationship types each visual question exhibits between its different answer grounding pairs, overall as well as with respect to each VQA source. Results are shown in Table \ref{table:num4Relationship}.  We find that most visual questions (i.e., 89\%) have just one relationship type between their answer groundings. We suspect it is because most of the visual questions only have two valid answers, two answer groundings, and thus one kind of relationship. When comparing results from the two VQA sources, we observe VQAv2 has slightly more relationships than VizWiz-VQA dataset. We suspect this is due to a more even percentage distribution across the four types of relationships we analyzed, as shown in Table~~\ref{table:4Relationship}. 

We next tally how many visual questions have each type of relationship, overall as well as with respect to each VQA source.  Results are shown in Table~\ref{table:4Relationship}.  Overall, we find VizWiz-VQA and VQAv2 have a similar distribution of answer grounding relationships. The most common relationship between answer groundings for a visual question is intersection, with this occurring for over half of the visual questions. This finding has important implications for both human visual perception and model development. We suspect that when multiple individuals provide different answers based on distinct visual evidence, they may be focusing on the \textit{same object} while paying attention to distinct \textit{details}, resulting in an intersection of visual evidence. 

\begin{table}[t!]
\centering
\begin{tabular}{llll}
\hline
 & All & VQAv2 & VizWiz-VQA                                             \\ \hline
1                        &  89\% (812)&  86\% (133) &   89\% (679)\\
2                        & 11\% (103)  &   14\% (22)  &   11\%  (81)                                      \\
3                        & 0\% (1)   & 0\%  (0)  &  0\% (1)                                               \\\hline
\end{tabular}   
\caption{Number of different kinds of relationships that a visual question's answers have, overall and per data source.  }
    \label{table:num4Relationship}
\end{table}

\begin{table}[t!]
\centering
\begin{tabular}{llll}
\hline
 & All & VQAv2 & VizWiz-VQA                                            \\ \hline
Disjoint                 &  10\% (99) &  16\% (28)   &  8\% (71) \\
Intersected              &  67\% (685) & 60\% (107)  &   68\% (578)                                             \\
Contained                & 15\% (151) &  12\% (21)  &  15\% (130)                                            \\
Equal                    & 8\% (86) & 12\% (21)  &  8\% (65)                                                \\ \hline
\end{tabular}        
\caption{Percentage of visual questions with multiple answer groundings having each relationship type between its answer groundings, overall and for each data source.}
    \label{table:4Relationship}

\end{table}

\begin{figure}[t!]
     \centering
     \includegraphics[width=0.47\textwidth]{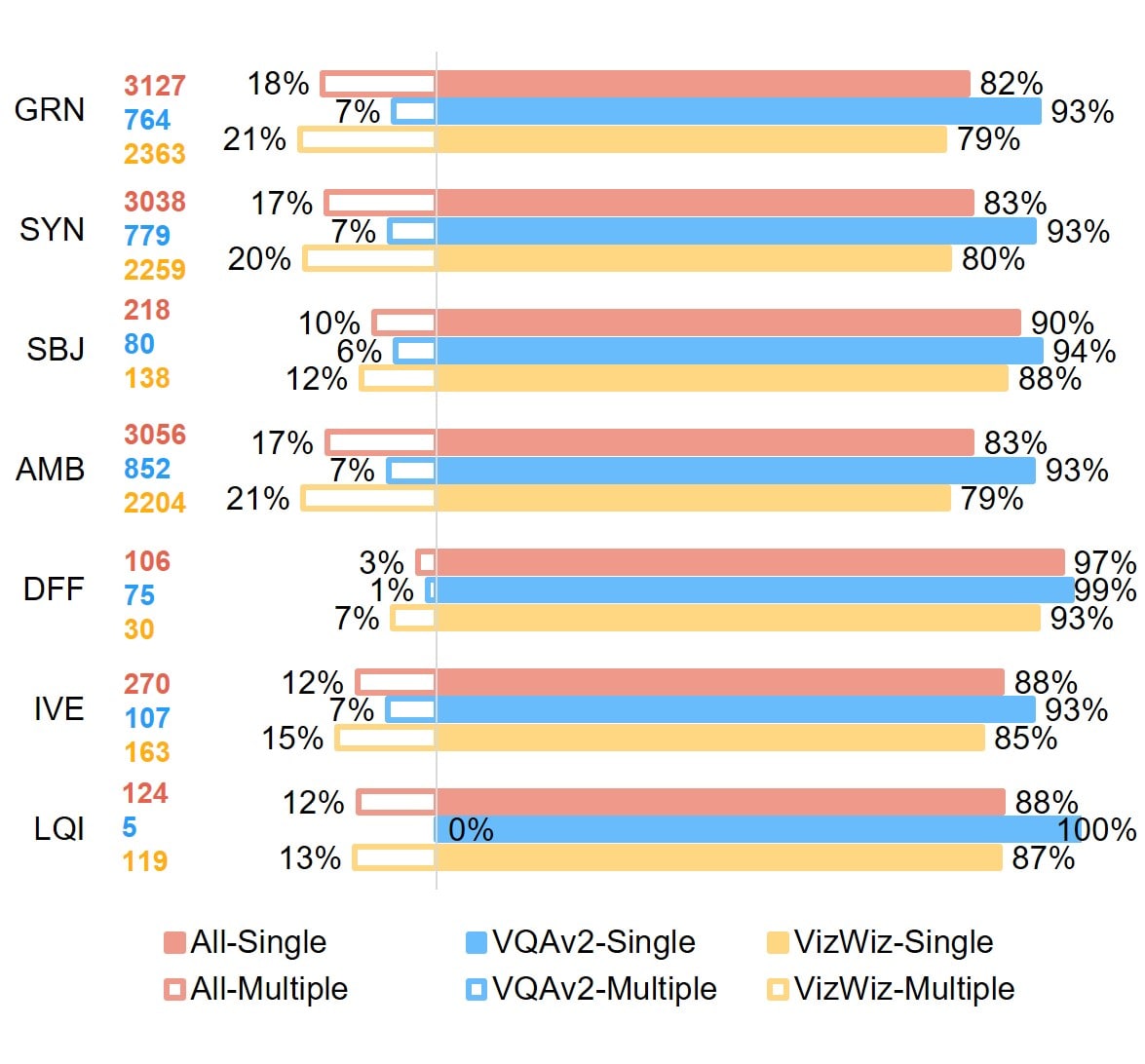}
     \vspace{-0.5em}
        \caption{Relationship of whether a visual question has a single grounding for all answers and reasons for different answers for the VQAv2 and VizWiz dataset sources.}
    \label{fig:7Relationship}
\end{figure}

\vspace{-1em}\paragraph{Relationship Between Why Answers Differ and Number of Answer Groundings.} 
We next analyze the tendency for visual questions that lead to a single versus multiple answer groundings to be associated with various reasons why natural language answers can differ.  For each visual question, we obtain the reasons why answers can differ using the following seven labels provided in the VQA-Answer-Difference dataset \cite{bhattacharya2019does}: low-quality image (LQI), insufficient visual evidence (INV), difficult questions (DFF), ambiguous questions (AMB), subjective questions (SBJ), synonymous answers (SYN), and varying levels of answer granularity (GRN).\footnote{We exclude the reasons ``Spam answer" and ``Invalid question" because, by definition, they cannot have grounded answers.}\textsuperscript{,}\footnote{As done in \cite{bhattacharya2019does}, we assign labels using a 2-person threshold.}  Results are shown in a bar chart in Figure \ref{fig:7Relationship}, with the left part showing percentages for visual questions that have multiple answer groundings and the right part showing percentages for visual questions with a single answer grounding.

\begin{figure}[t!]
     \centering
     \includegraphics[width=0.5\textwidth]{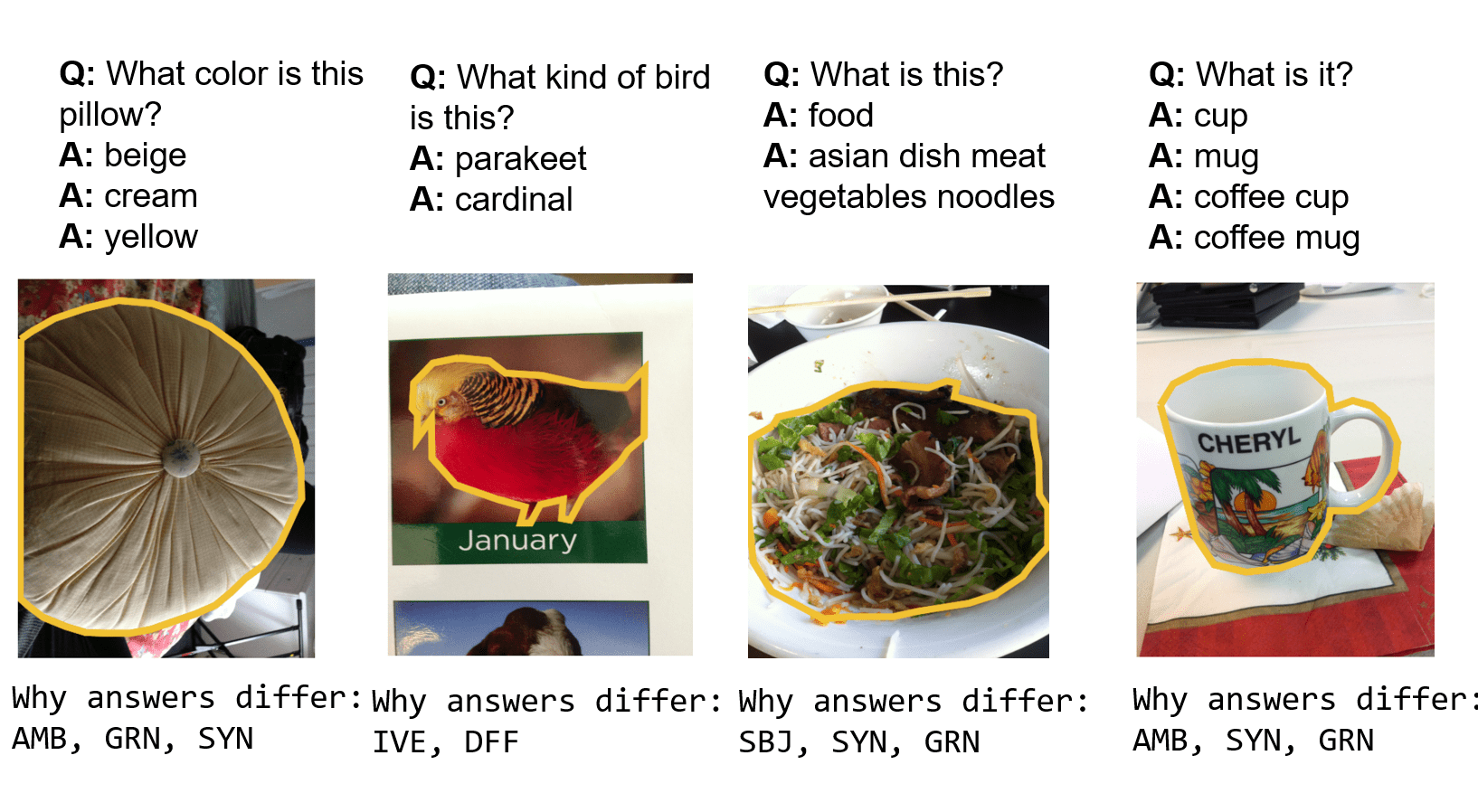}
     \vspace{-2em}
        \caption{Visual questions with one answer grounding alongside annotations indicating why answers differ \cite{bhattacharya2019does}.}
    \label{fig:samegroundingExample}
\end{figure}

Overall, visual questions with multiple answer groundings commonly are associated with varying levels of answer granularity (GRN), ambiguous questions (AMB), and synonymous answers (SYN). The nearly identical results for AMB and SYN are not surprising since over 85\% of VQAs labeled as SYN also occur with AMB in both the VizWiz-VQA and VQAv2 datasets.  

Visual questions labeled with difficult (DFF) tend to share a single grounding. Intuitively, this makes sense as there is consensus around what the question is asking about but people simply struggle to know what is the correct answer.  An example of this scenario is shown in Figure \ref{fig:samegroundingExample}, with the question ``what kind of bird is this?"

When comparing results from the two VQA sources, we observe that VQAv2 and VizWiz-VQA have large differences (larger than 10\%) for four reasons: GRN, SYN, AMB, and LQI. Examples of visual questions that are labeled as AMB, SYN, GRN are exemplified in Figure \ref{fig:samegroundingExample} (col 1, 3, and 4).

\begin{figure}[t!]
     \centering
     \includegraphics[width=0.4\textwidth]{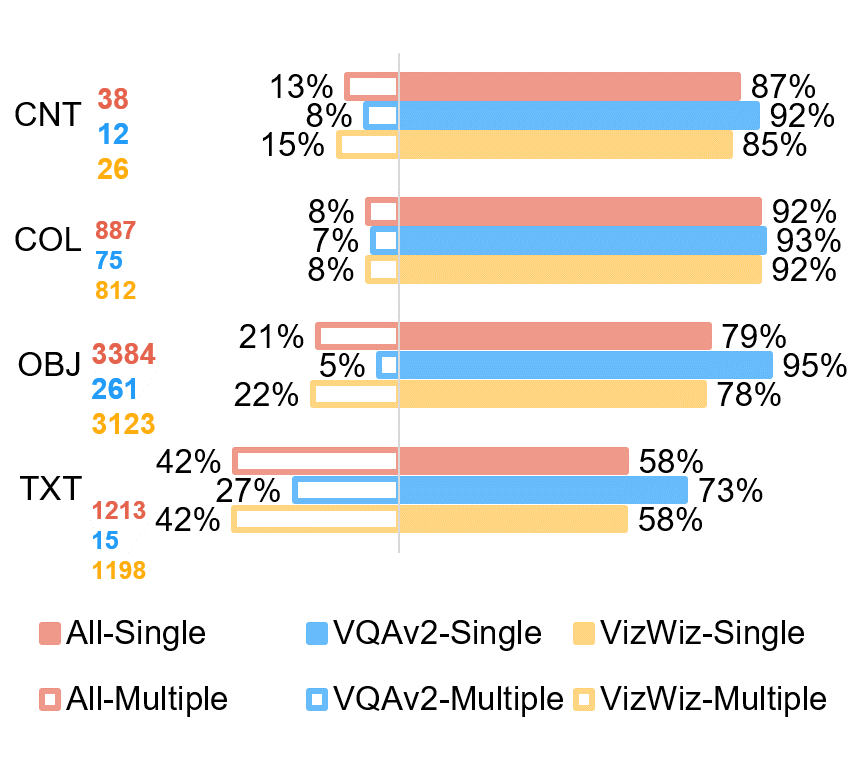}
     \vspace{-1em}
        \caption{Amount of multiple answer groundings per visual question for four vision skills, overall and per dataset source (VQAv2 and VizWiz-VQA). }
    \label{fig:samediffvisionskills}
\end{figure}

\vspace{-1em}\paragraph{Relationships Between Vision Skills Needed to Answer a Visual Question and Number of Answer Groundings.}
We next evaluate how the vision skills needed to answer a visual question relate to whether a visual question has a single grounding.  The following labels for the four vision skills are provided in the VizWiz-VQA-Skills dataset \cite{zeng2020vision}: object recognition (OBJ), text recognition (TXT), color recognition (COL), and counting (CNT). Following \cite{zeng2020vision} we use majority vote from the 5 annotations to determine the vision skill labels.  We perform our analysis over all visual questions as well as with respect to each VQA source independently.   Results are shown in Figure \ref{fig:samediffvisionskills}, based on observed percentages for each VQA source. 

Overall, we found that visual questions trying to read \textit{text} tend to have multiple answer groundings. One common example is visual questions about products, as exemplified in Figure \ref{fig:main} (e.g., chips product).  In contrast, questions related to recognizing \textit{color} tend to have a single answer grounding. We suspect people might express `color' in different ways because of individual or cultural differences, despite often looking at the same region. For example, a question asking ``What is the color of this cloth?" might get different of answers ``khaki", ``tan", and ``brown" despite all referring to the same region (i.e., the cloth). 

\begin{figure}[t!]
     \centering
     \includegraphics[width=0.5\textwidth]{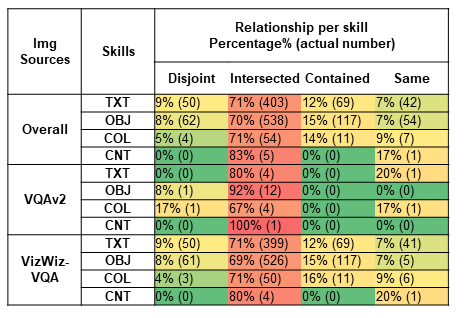}
     \vspace{-2em}
        \caption{The heatmap table shows the percentage and the number of relationships between answer groundings with respect to each of the four vision skills for our dataset (overall) and for each image source (VQAv2 and VizWiz-VQA). }
    \label{fig:answergroundingsVisonSkillsRelationship}
\end{figure}

We also evaluate relationships between visual questions that result in multiple answer groundings with respect to each of the four vision skills overall as well as with respect to each VQA source independently. We determine if a visual question has a single grounding and what skills are needed following the same process of the previous analysis. Results are shown in Figure \ref{fig:answergroundingsVisonSkillsRelationship}.\footnote{Results for VQAv2 may not be representative since only a small amount of our dataset's visual questions have vision skill labels.}  Overall, we observe that visual questions related to ``text recognition" and ``object recognition" are more likely to have a ``disjoint" relationship compared to ``color recognition" and ``counting" skills. Examples are shown in Figure \ref{fig:main} (``cabinets" and ``hood" are disjoint) and Figure \ref{fig:highquality} (``blanket" and ``umbrella" are disjoint; ``rsb" and ``royal society for blind" are disjoint).
\section{Algorithm Benchmarking}
\label{section:algorithm}
Using the VQA-AnswerTherapy dataset, we now quantify how well modern architectures support two novel tasks of (1) predicting if a visual question shares the same grounding for all answers and (2) localizing all groundings for all answers to a visual question.

\vspace{-1em}\paragraph{Dataset Splits.}  
Our VQA-AnswerTherapy dataset contains  3,794, 646, and 1,385 for train/val/test sets, respectively. The visual questions from the VizWiz-VQA dataset are split to match the train/val/test splits of the original VizWiz-VQA dataset \cite{gurari2018vizwiz}. Our dataset also has visual questions originating from the training set of the VQAv2 dataset \cite{balanced_vqa_v2}, which is split into train/val/test splits using 70\%, 10\%, and 20\% of the data respectively.

\begin{table*}[t!]
  \centering
        \begin{tabular}{ l  c  c  c c | c c c  c }
    \toprule
      & \multicolumn{4}{c}{{\bf Precision}} & \multicolumn{4}{c}{{\bf Recall}}  \\ 
    \cmidrule(r){2-5} \cmidrule(r){6-9} 
     \textbf{Model Type:}
    & {\small \textit{ViLT}}
    & {\small \textit{mPLUG-Owl}}
    & {\small \textit{ Na\"ive (M) }}
    & {\small \textit{ Na\"ive (S) }}

    & {\small \textit{ViLT}}
    & {\small \textit{mPLUG-Owl}}
    & {\small \textit{ Na\"ive (M) }}
    & {\small \textit{ Na\"ive (S) }} \\
    \midrule
       \textbf{All:S} &  \textbf{0.86} & 0.80 & - & 0.80   &0.94 &0.82& 0.0 & \textbf{1.0 } \\ 
       \textbf{VQAv2:S}  & \textbf{0.93}   &  \textbf{0.93} & - & 0.92& 0.97 &  0.81& 0.0&\textbf{1.0}\\ 
       \textbf{VizWiz-VQA:S} & \textbf{0.82}  & 0.74  & - & 0.74&0.92   & 0.83  & 0.0&\textbf{1.0 } \\ 
       \hdashline
       \textbf{All:M}  &  \textbf{0.59} & 0.20  & 0.20 &  -&0.37&0.20  & \textbf{1.0 } & 0.0 \\ 
       \textbf{VQAv2:M} & \textbf{0.24}  &0.11  &0.08  & - &0.11&0.26   &\textbf{1.0 } &0.0 \\ 
       \textbf{VizWiz-VQA:M}  & \textbf{0.63}&0.25   &0.26 & -  &0.42  &0.17&\textbf{1.0 } &0.0\\ 
       \bottomrule	 
  \end{tabular}
        \caption{Performance of methods at predicting whether a visual question will result in answers that all share single (i.e., `S') or multiple (i.e., `M') groundings respectively, overall as well as with respect to each data source.  Of note, no values (`-') are entered for some models because they do not yield valid scores.  This includes the Na\"ive (M) for task `S' with respect to precision and Na\"ive (S) for task `M' with respect to precision, because no positives are predicted, making the denominator zero (i.e., \text{precision} = \( \text{True Positive} / (\text{True Positive} + \text{False Positive}) \). This also includes the Na\"ive (M) model for task `S' with respect to recall and Na\"ive (S) model for task `M' with respect to recall, because there are no positives and so the numerator is again zero (i.e., \text{recall} = \text{True Positive} / (\text{True Positive} + \text{False Negative}).}
        ~\label{table_singleAnswerGrounding}
\end{table*} 

\subsection{Single Answer Grounding Challenge}
The task is to predict if a visual question will result in answers that all share the same grounding. For completeness, we also explore the predicting if a visual question will result in answers that multiple groundings.  We evaluate methods using two standard metrics for binary classification tasks: precision and recall.

\vspace{-1em}\paragraph{Models.} 
We benchmark four models.  We fine-tune a top performing algorithm for the VQA task, ViLT \cite{kim2021ViLT}, on the training set on of our entire dataset. To do so, we modified the output layer of the architecture with a two-class softmax activation to support binary classification. We also benchmark the state-of-the-art vision and language foundation model which was the first to achieve human parity on the VQA Challenge,  mPLUG-Owl \cite{ye2023mplug}, in a zero-shot setting with the prompt ``Do all plausible answers to the question indicate the same visual content in this image?" This zero-shot setting is a useful baseline because of the imbalanced and relatively small size of our dataset to support training.  We finally benchmark two na\"ive baselines that each only predict one label, i.e., all samples share the same answer grounding \emph{or} all samples have multiple answer groundings.

\begin{figure}[t!]
     \centering
     \includegraphics[width=0.5\textwidth]{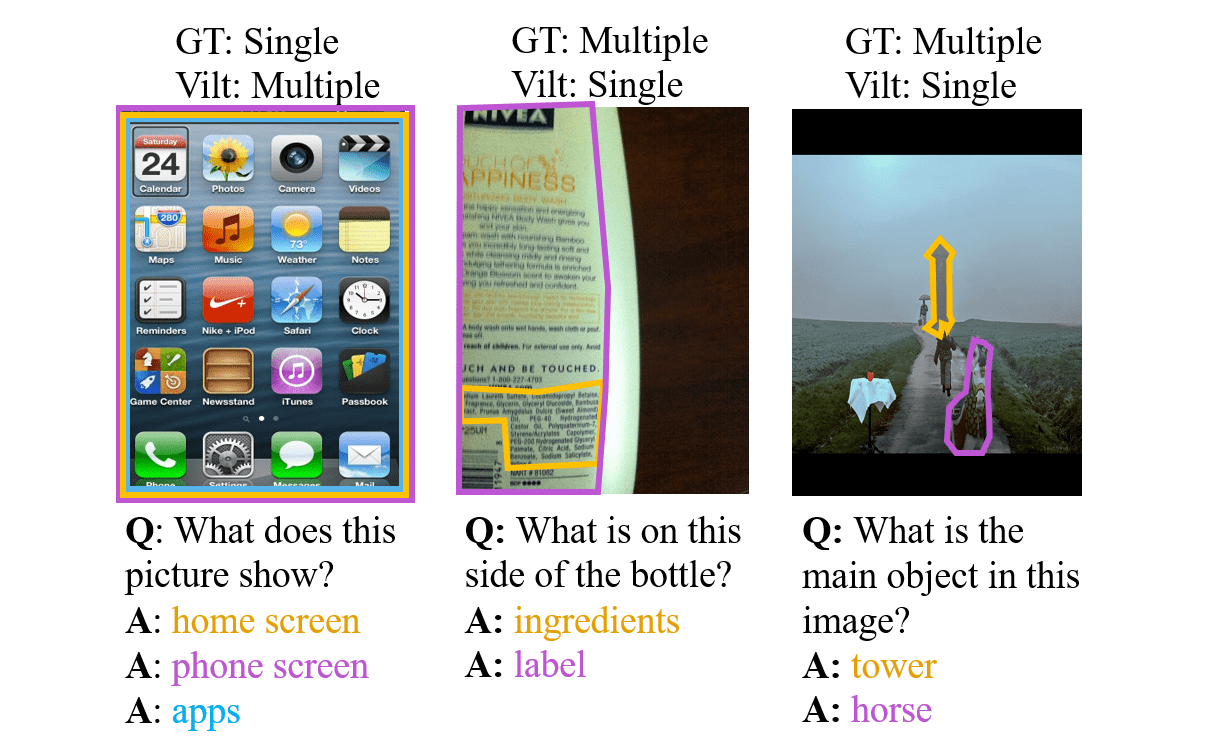}
     \vspace{-1em}
        \caption{Qualitative results for the fine-tuned ViLT model on the Single Answer Grounding Challenge, alongside the question-answer pair and ground truth answer groundings.}
    \label{fig:SingleGroundinChallengeQualitative}
\end{figure}

\vspace{-1em}\paragraph{Results.}
Results are reported in Table \ref{table_singleAnswerGrounding} for predicting whether a visual question has answers that all share single grounding (`S') and predicting whether a visual question has answers with multiple groundings (`M').   Testing results are reported for the entire dataset, as well as on each VQA source (VQAv2 and VizWiz-VQA) independently.  

We observe that it is much more difficult to predict when answers have multiple groundings compared to a single grounding, i.e., both ViLT and mPLUG-Owl receive much lower precision and recall when predicting whether there are multiple answers grounding compared to predicting if there is a single answer grounding. This is true both for the fine-tuned ViLT model, which is susceptible to failing from the data imbalance (i.e., only 15.7\% of visual questions have multiple answers grounding), as well as the the zero-shot mPLUG-Owl solution.  

We next analyze overall performance for the models.  While ViLT is an inferior VQA model compared to mPLUG-Owl, it achieves better performance once fine-tuned on our dataset for our novel task. This enhancement is striking, considering the limited size of our dataset. This observation underscores the value of our dataset, illustrating how even a modest number of samples can bolster the robustness of current models.  In contrast, the foundation model, mPLUG-Owl, achieves inferior or comparable performance to a na\"ive baseline. We manually inspected all examples where the top-performing fine-tuned ViLT struggles, and show examples in Figure \ref{fig:SingleGroundinChallengeQualitative}. For instances where there is a single answer grounding and ViLT predicts multiple answer groundings, across both VizWiz-VQA and VQAv2 sources, often images show text or multiple objects while the question typically references the entire object or a particular area, as illustrated in Figure \ref{fig:SingleGroundinChallengeQualitative} column 1. Conversely, in situations where there are multiple answer groundings but ViLT predicts only one (143 examples for VizWiz-VQA and 34 for VQAv2), we observe distinct patterns between the VizWiz-VQA and VQAv2 datasets. Specifically, in the VizWiz-VQA dataset, 98 out of the 143 instances occur because the image contains only one significant object, with the answer primarily focusing on text recognition (Figure \ref{fig:SingleGroundinChallengeQualitative} column 2). In contrast, for the VQAv2 dataset, this discrepancy arises in 27 out of the 34 cases mainly because the question is ambiguous about which object/area it is asking about and the image contains multiple objects (Figure \ref{fig:SingleGroundinChallengeQualitative} column 3).

When comparing the performance across datasets, despite that we permitted both models to have an unfair advantage that they could observe during training the COCO images that are used in the VQAv2 dataset \footnote{ViLT is pretrained on GCC, SBU, COCO, and VG datasets and mPLUG-Owl is pretrained on LAION-400M, COYO-700M, Conceptual Captions and MSCOCO.} and it's cheating to test it on the training set of the VQAv2 dataset, we only observe higher performance when predicting ``VQAv2:S" compared to ``VizWiz-VQA:S" and didn't observe higher performance when predicting ``VQAv2:M" compared to ``VizWiz-VQA:M". We suspect the reason is that the VQA-Single Answer Grounding dataset is highly imbalanced with 93\% of visual questions having different answers that all describe the same visual evidence. 

\subsection{Answer(s) Grounding Challenge}
Given an image and a question, the task is to predict the image region to which the answer is referring.

\vspace{-1em}\paragraph{Evaluation Metric.} We measure the similarity of each binary segmentation to the ground truth with IoU. We report two IoU scores, IoU and IoU-PQ (IoU-Per Question). The IoU-PQ uses as the score for each visual question the average of the IoU scores for all answer groundings to that visual question. We utilize IoU-PQ in place of IoU because the metadata (e.g., single/multiple annotations, vision skills annotations) for fine-grained analysis pertains to each visual question rather than each answer grounding.

\vspace{-1em}\paragraph{Models.}
We evaluate three variants for each of the following three models: SeqTR, UNINEXT \cite{yan2023universal}, and SEEM  \cite{zou2023segment}.\footnote{We do not benchmark answer grounding models~\cite{wang2022unifying,zhang2019interpretable,urooj2021found} since these show weak performance on existing challenges (e.g., \cite{chen2022grounding}.)} For the three variants per model, we feed the model the image-question pair (i.e., Model(I+Q)), the image-question-answer triplet (i.e., Model(I+Q+A)) and the image-answer pair (i.e., Model(I+A). We fine-tuned a top-performing referring segmentation algorithm, SeqTR, on our entire dataset. SeqTR \cite{zhu2022seqtr} is pretrained on a large corpus of datasets (i.e., \cite{krishna2017visual,yu2016modeling,mao2016generation,kazemzadeh2014referitgame,plummer2015flickr30k,nagaraja2016modeling} ).  We also evaluated zero-shot performance for both the UNINEXT \cite{yan2023universal} and SEEM \cite{zou2023segment} models. We selected UNINEXT because of its state-of-the-art performance for the Referring Expression Segmentation task and SEEM since it claims to “segment everything everywhere”.

\begin{table}[b!]
\centering
\begin{tabular}{cccc}
\hline
\textbf{Models}  &\textbf{All} & \textbf{VQAv2} & \textbf{VizWiz-VQA}\\\hline

SeqTR (I+Q+A)   &   \textbf{66.68} & \textbf{64.50}   & \textbf{ 67.89} \\
SeqTR (I+Q) & 62.04 &  58.46 &   64.02 \\
SeqTR (I+A) &  63.27  &   58.03 &   66.17 \\\midrule
SEEM (I+Q+A)          &  \textbf{53.77}    & \textbf{50.67} &    \textbf{55.49} \\
SEEM (I+Q)          &     45.17   &  44.65   & 45.46   \\
SEEM (I+A)          &   52.10   &  46.83   &  55.03  \\\midrule
UNINEXT (I+Q+A) & \textbf{53.76} &\textbf{42.73}& \textbf{59.88} \\
UNINEXT (I+Q) & 50.51 & 40.96 & 55.81 \\
UNINEXT (I+A)& 52.76 & 41.60 & 58.95\\

\hline

\end{tabular}
\vspace{-0.25em}\caption{Performance of models for predicting all answer groundings per visual question.}
\label{table:SeqTR results}
\end{table}

\vspace{-1em}\paragraph{Overall Results.}
Results are shown in Table \ref{table:SeqTR results}.\footnote{Due to space constraints, we report overall model performance with respect to IoU-PQ in the supplementary materials. The scores align closely with IoU.} Performance is reported for the entire dataset (column 2) as well as with respect to each VQA source independently (column 3 and column 4). 

As shown, all analyzed models perform poorly. For example, the top-performing SeqTR(I+Q+A) overall only achieves an IoU of 66.68\%.  This arises despite that all three models were exposed to COCO images in the pretraining phases; performance on VQAv2 dataset is still similar to that for VizWiz-VQA. We suspect the referring segmentation pretraining may result in models taking shortcuts by remembering images while ignoring the language (the language inputs when pretraining are referring expressions, which can differ considerably from our inputs).

\begin{figure*}[t!]
     \centering
     \includegraphics[width=1\textwidth]{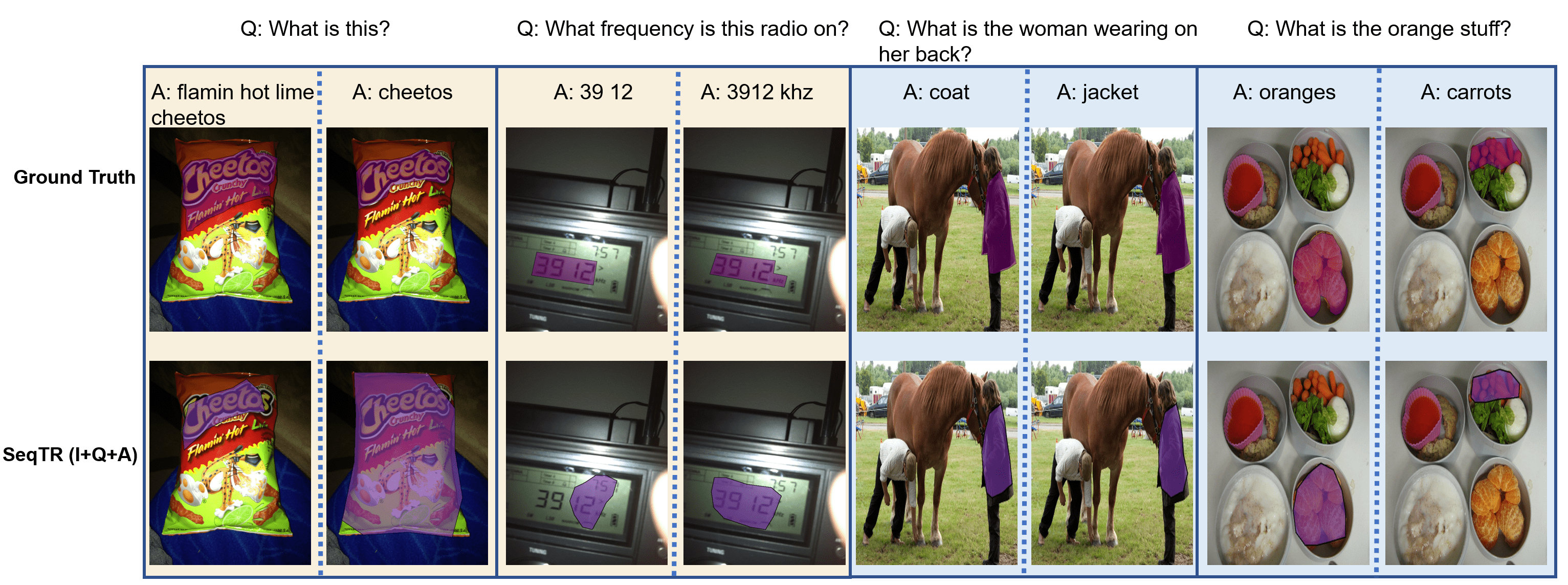}
     \vspace{-2.3em}
        \caption{Qualitative results from SeqTR (I+Q+A) for visual questions coming from VizWiz-VQA (yellow background) and VQAv2 (blue background).}
    \label{fig:algorithm_example}
\end{figure*}

We also analyze the results for each model. While part of the poor performance of SeqTR could be attributed to the relatively small amount of training examples available for fine-tuning, our results in Table \ref{Table:samediffSeq} offer strong evidence that the challenge of grounding different answers is also an important factor.  That is because SeqTR(I+Q+A) scores 72\% on visual questions with a single answer grounding versus 43.66\% on visual questions with multiple answer groundings, underscoring a greater difficulty for the latter task.  Our results on UNINEXT and SEEM also underscore how current large segmentation models lack sufficient zero-shot generalization capabilities, a necessary prerequisite for applications such as open-domain VQA. 

Comparing the performance across different variant settings (I+Q+A/I+Q/I+A), we find that the model that receives the most information as input (I+Q+A) performs best, which aligns with our intuition. We show the qualitative results for SeqTR (I+Q+A) model in Figure 
\ref{fig:algorithm_example}. We observed models often fail for vision questions with multiple answer groundings that require recognizing text. In contrast, models often perform well for visual questions that identify common objects.

\begin{table}[t!]
\centering
{
\begin{tabular}{lcccc}

\hline
\textbf{Models}&
\textbf{Single}  & \textbf{Multiple}  \\ \toprule

\textbf{SeqTR (All)}   &  \textbf{71.69} & \textbf{43.66}    \\
\textbf{SEEM (All)}   &  60.45  & 22.75     \\
\textbf{UNINEXT (All)}   & 60.47   & 23.08     \\   \midrule

\textbf{SeqTR (VQAv2)}   &  \textbf{65.56} &  \textbf{49.66} \\ 
\textbf{SEEM (VQAv2)}   & 51.65   &  31.70  \\  
\textbf{UNINEXT (VQAv2)}   & 43.79 & 24.15    \\ \midrule
\textbf{SeqTR (VizWiz-VQA)}  & \textbf{75.94} & \textbf{42.66}   \\
\textbf{SEEM (VizWiz-VQA)}   & 66.56   & 21.27 \\
\textbf{UNINEXT (VizWiz-VQA)}  & 72.06& 22.90   \\
\bottomrule 
\end{tabular}}
\caption{Performance of models at localizing all answer groundings with respect to IoU-PQ scores. They struggle most for visual questions with multiple answer groundings.}
    \label{Table:samediffSeq}
\end{table}

\vspace{-1em}\paragraph{Analysis With Respect to Single vs Multiple Answer Groundings.}
Table \ref{Table:samediffSeq} presents the IoU-PQ scores for visual questions with respect to visual questions with a single answer grounding and multiple answer groundings. We use the settings of (I+Q+A) for each model to reveal the upper bound of what is possible from top-performing models.  

We observe that the top-performing model, SeqTR, largely lacks the ability to predict multiple answer groundings.  This suggests modern models are designed based on an incorrect assumption that only one answer grounding is needed for a visual question.  Still, SeqTR significantly outperforms SEEM (All) and UNINEXT (All), highlighting a potential benefit of a modest amount for fine-tuning models for our target task.

Delving into the data based on VQA sources, a compelling pattern emerges. All models consistently deliver superior performance for visual questions with multiple answers groundings on VQAv2 compared to VizWiz-VQA. Conversely, performance for visual questions with a single answer grounding is worse on VQAv2 than for VizWiz. One potential factor leading to this outcome may stem from VQAv2 having a higher prevalence of complex scenes and so presenting a greater difficulty for grounding answers when only a single grounding is needed. 
\section{Conclusions}
This work acknowledges a fundamental challenge that visual questions can have multiple valid answers. We support further exploration of this fact by introducing a new dataset, which we call VQA-AnswerTherapy, that provides a grounding for every valid answer to each visual question.  We also propose two novel challenges of (1) predicting whether a visual question has a single answer grounding (versus multiple answer groundings) and (2) locating all answer groundings for a given visual question.  Our algorithm benchmarking results reveal that modern methods perform poorly for these tasks, especially when a visual question has multiple answer groundings.  We share our dataset and crowdsourcing source code to facilitate future extensions of this work.

\paragraph{Acknowledgments.}
This work was supported with funding from Microsoft AI4A and Amazon Mechanical Turk.  

\clearpage

{\small
\bibliographystyle{ieee_fullname}
\bibliography{egbib}
}
\clearpage

\section{Supplementary Material}
This document supplements the main paper with more information about:
\begin{itemize}
      \item Dataset collection (Supplements Section 3.1) 
    
    \begin{itemize}
            \item Method for hiring expert crowdworkers (Supplements Section 3.1)
            \item Annotation task interface (Supplements Section 3.1)
            \item Method for reviewing work from crowdworkers (Supplements Section 3.1)
    \end{itemize}
    
    \item Dataset analysis (Supplements Section 3.2)
        \begin{itemize}
        \item Incorrect answer (Supplements Section 3.2)        
        \item No polygon and multiple polygons (Supplements Section 3.2)   
        \item Grounding agreement (Supplements Section 3.2)            
        \item Reconciling redundant annotations (Supplements Section 3.2)
        \item Four grounding relationships (Supplements Section 3.2) 
        \item Most common answers (Supplements Section 3.2)
    \end{itemize}
    \item Algorithm benchmarking (Supplements Section 4)
        \begin{itemize}
        \item Experimental details for Single Answer Grounding Challenge (Supplements Section 4.1)
        \item Experimental details for Answer(s) Grounding Challenge (Supplements Section 4.2)
        \item Performance of three models for Answer(s) Grounding Challenge with IoU-PQ metric (Supplements Section 4.2)     
        \item Answer(s) Grounding Challenge: qualitative results for model benchmarking (Supplements Section 4.2)                
    \end{itemize}
\end{itemize}

\renewcommand\thesection{\Roman{section}}
\setcounter{section}{0}

\section{Dataset Collection}
\subsection{Method for Hiring Expert Crowd Workers.}
We hired 20 workers who completed our one-on-one zoom training, passed our multiple qualification criteria, and consistently generated high-quality results. We limited the number of workers on our task to prioritize collecting \emph{high-quality} annotations over the \emph{efficiency} that would come with having more workers; i.e., it is easier to track the performance of fewer workers. We gave our 20 workers our contact information so that they could send any questions about the tasks and receive feedback quickly. 

We paid above the US federal minimum wage to simultaneously support ethical data collection and encourage workers to create higher-quality results.  Our average hourly wage was 9.64 dollars/hour.  This rate is derived using the median time it took to annotate the 1,000 HITs collected in our pilot study (i.e., 2.49 minutes per HIT) with the amount we paid per HIT (i.e., 0.4 dollars/HIT).

\subsection{Annotation Task Interface.}
\begin{figure*}
         \centering
     \includegraphics[width=0.8\textwidth]{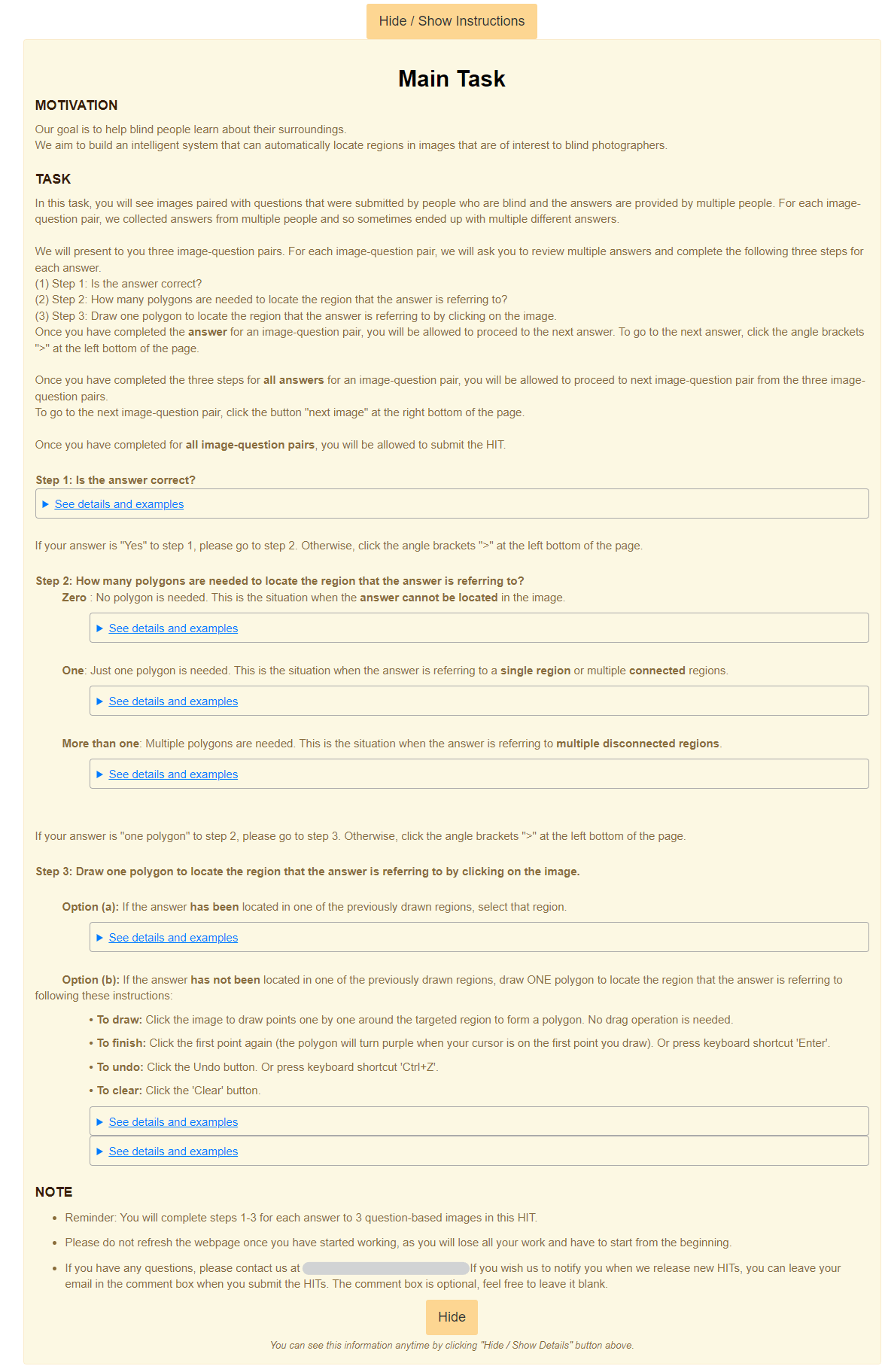}
        \caption{Instructions for our annotation task. }
        \label{fig:instruction}
\end{figure*}

\begin{figure*}
     \centering
     \begin{subfigure}[b]{0.9\textwidth}
         \centering
     \includegraphics[width=0.9\textwidth]{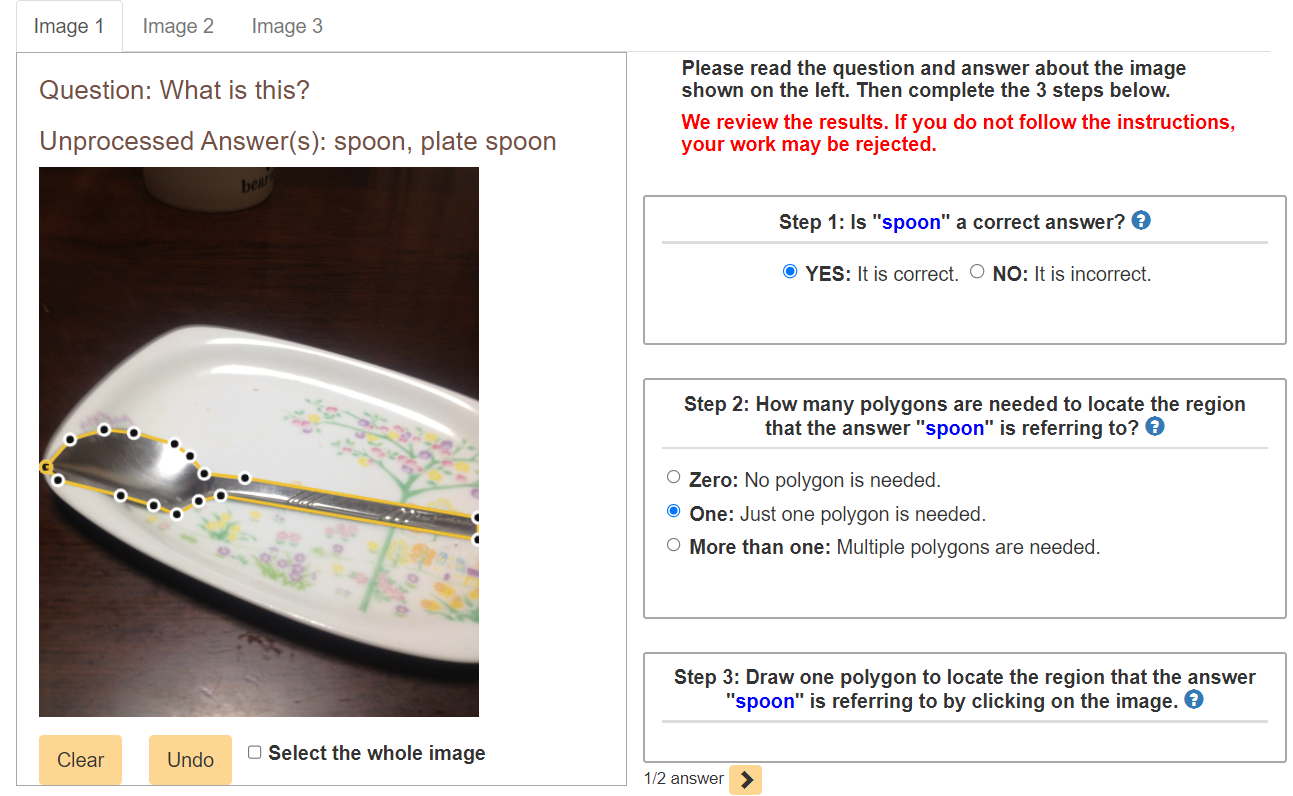}
         \caption{User interface to ground the different answers for each visual question.}
         \label{fig:y equals x}
     \end{subfigure}
     \hfill
     \begin{subfigure}[b]{0.9\textwidth}
         \centering
     \includegraphics[width=0.94\textwidth]{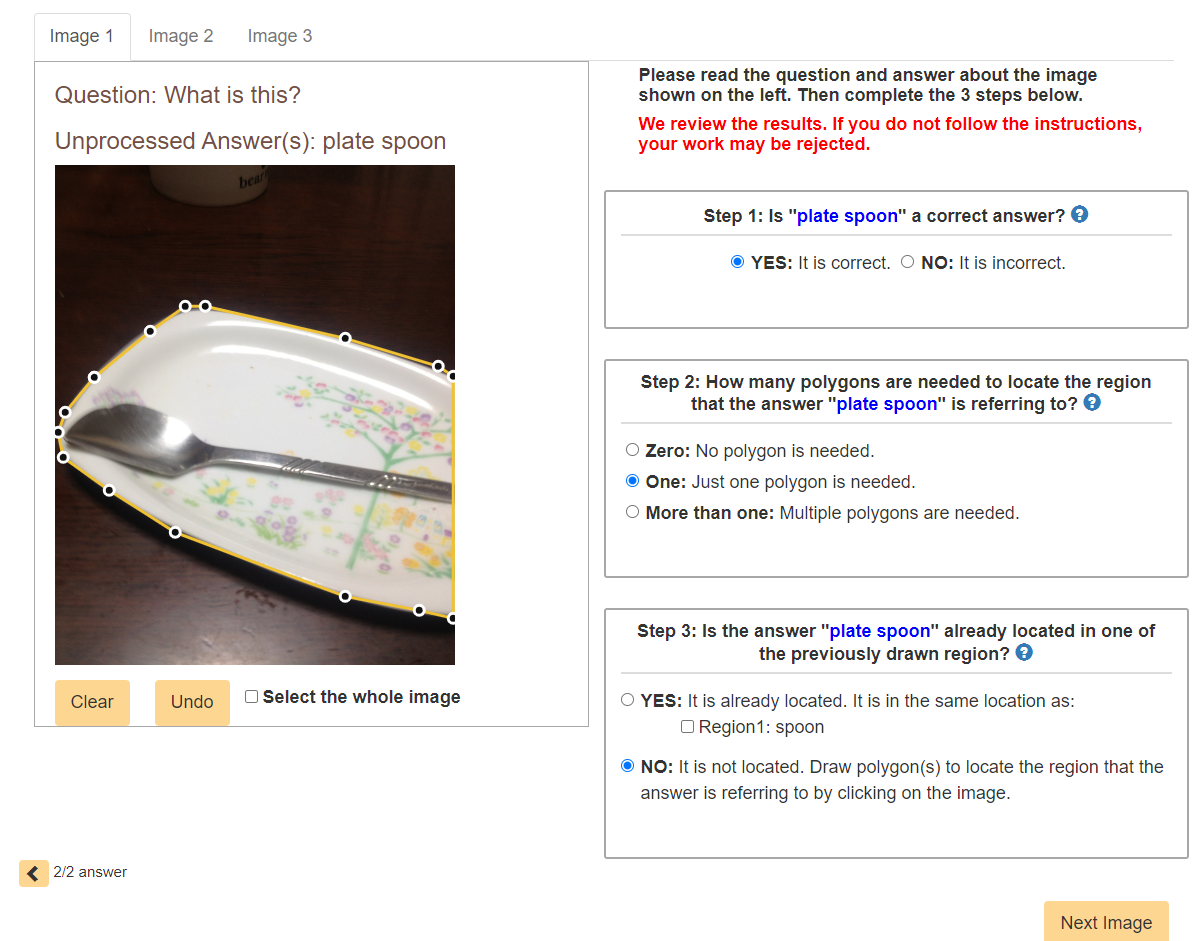}
         \caption{After one answer grounding was available for a visual question, the annotator could choose between selecting a previously drawn polygon as the grounding for the new answer and drawing a new polygon to ground the answer.
}
         \label{fig:userinterface2}
     \end{subfigure}
        \caption{Screenshots of our annotation task interface. }
        \label{fig:userinterface}
\end{figure*}

We show a screenshot of the crowdsourcing instructions in Figure \ref{fig:instruction} and the interface to collect annotations in Figure \ref{fig:userinterface}. The link to this code is available at https://github.com/CCYChongyanChen/VQATherapyCrowdsourcing/.

\begin{figure*}[tph]
     \centering
     \includegraphics[width=0.97\textwidth]{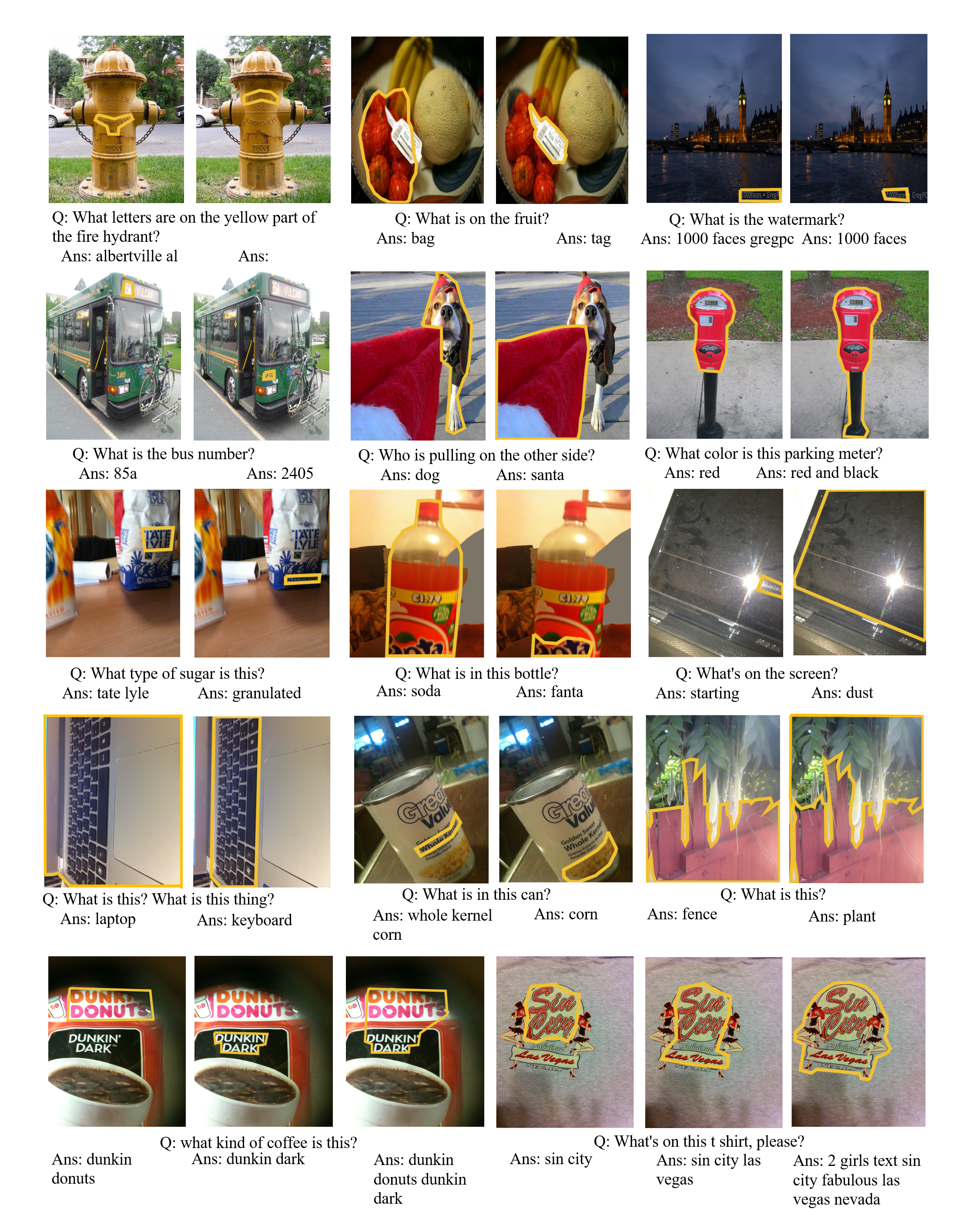}
     \vspace{-1.5em}
        \caption{High-quality grounding annotations for visual questions where valid answers refer to different groundings. The first two rows of examples come from VQAv2 dataset and the last three rows of examples come from VizWiz-VQA dataset.  }
    \label{fig:diffhighqualityexamples}
\end{figure*}

\begin{figure*}[tph]
     \centering
     \includegraphics[width=0.92\textwidth]{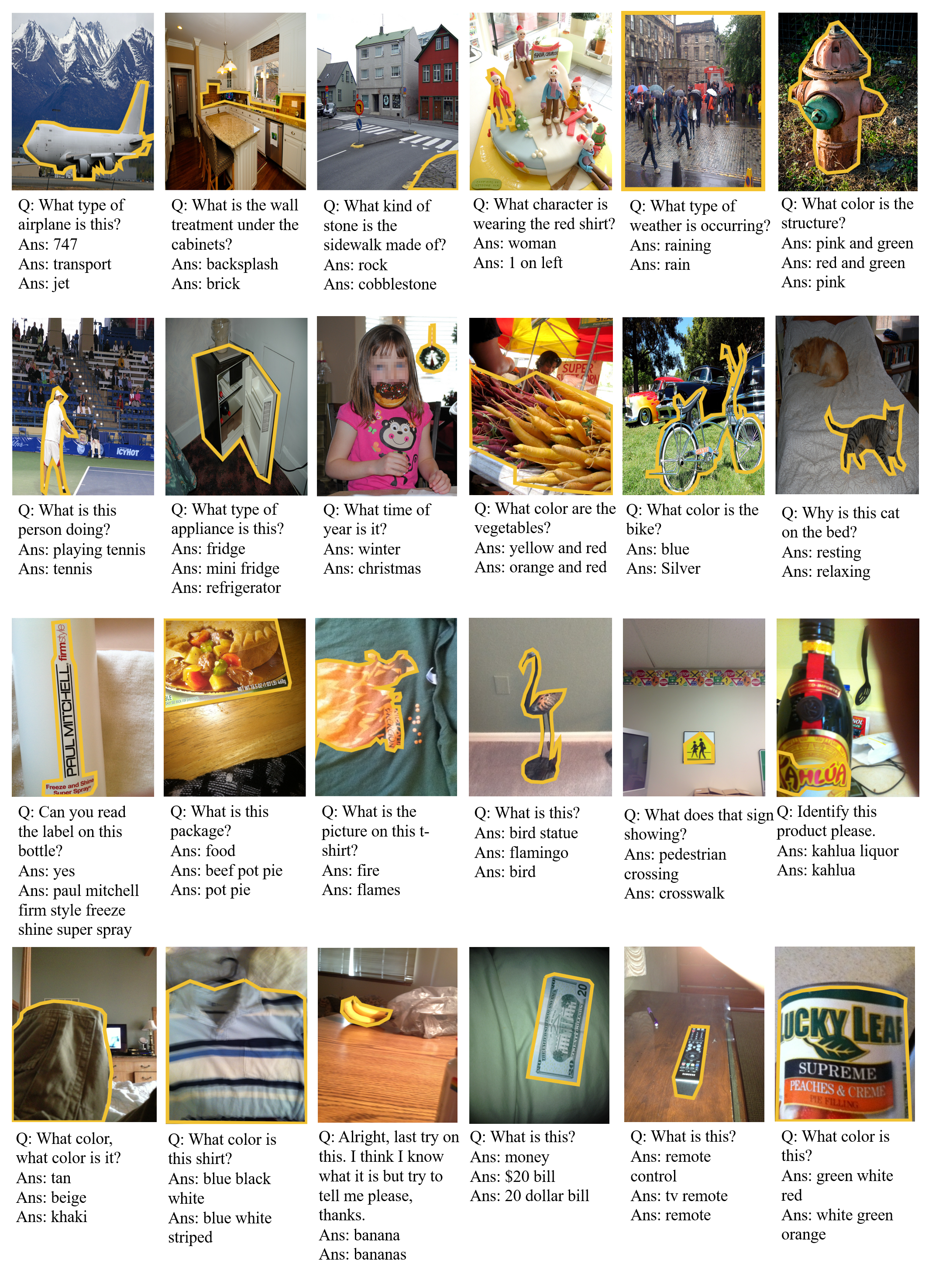}
      \vspace{-1.5em}
        \caption{High-quality grounding annotations for visual questions where all valid answers refer to the same grounding. The first two rows of examples come from VQAv2 dataset and the last two rows of examples come from VizWiz-VQA dataset. }
    \label{fig:samehighqualityexamples}
\end{figure*}

\subsection{Method for Reviewing Work from Crowdworkers.}
In the first three days crowdworkers worked for us\footnote{The data collection process lasted for 26 days.}, we conducted highly interactive quality control.  We conducted at least three inspections for each worker and gave them feedback continually. Each time, we viewed ten random HITs from each worker, provided each worker feedback if needed, and answered any questions by email or zoom. After the first time of review, 12 out of 20 workers passed our inspection without any issues. After the second time of review, 18 out of 20 workers passed our inspection without any issues. After the third time of review, all 20 workers demonstrated mastery of our task. We continued to monitor work from the eight workers who didn't work perfectly in the first time to ensure high-quality results.

As data collection proceeded, we leveraged a combination of automated and manual quality control steps to ensure the ongoing collection of high-quality results.  For automated quality control, we calculated the mean number of times each worker selected ``No" in Step 1 (contains incorrect answer), ``Zero" and ``More than one" in Step 2 (needs no polygon or more than one polygon) per HIT for each worker. If the mean was more than 1.25 times the mean value we observed across all workers, we randomly inspected at least ten HITs from that worker's recent submissions. We also monitored the mean time each worker spent on each HIT. When the mean was less than 1 minute, we randomly inspected at least ten HITs from this worker's recent submissions. Finally, we also monitored the mean of the number of points for an image (if applicable) drawn by each worker. When it was less than five points, we randomly inspected at least ten HITs from this worker's recent submissions and provided feedback as needed. For manual quality control, we continuously reviewed random selections of submitted HITs and provided feedback, when necessary, to workers throughout the data collection process (though after the first week, we hardly noticed any issues).

Statistics for each step of filtration are shown in Table~\ref{tab-dataset-statistics} to complement statistics provided in the main paper. Note that we only start crowdsourcing from a fraction of VQAv2's training set with 9,213 overlapping with [3] and 9,000 randomly sampled.

\begin{table}[h!]
\centering
\small
\begin{tabular}{cccc}
\hline
      & VizWiz & VQAv2 training& All \\ \hline
Original datatset                                             &  32,842      &  443,757 &   476,599   \\ 
Valid Answers                                             &   9,810    &  164,757 & 174,567 \\
Sub-questions                                                 &   9,528    &  163,731 &   173,259 \\ 

Crowdsourcing                                               &   9,528    &  [Sampled] 18,213 &27,741 \\

Incorrect Answers         &   7,216    &  8,214 &  15,430\\
No/multi polygons           &   6,729   &  5,561 &  12,290\\
75\% agreement           &   3,442    &  2,383 &  5,825\\\hline
\end{tabular}
\caption{Number of visual questions left after each step. We filtered visual questions with less than one valid answer/answer grounding after each step if applicable. }
\label{tab-dataset-statistics}
\end{table}

Examples of high-quality answer grounding results are shown in Figures \ref{fig:diffhighqualityexamples} and \ref{fig:samehighqualityexamples}. Figure \ref{fig:diffhighqualityexamples} shows visual questions that require \textit{text recognition} skill tend to have different groundings for all valid answers to a visual question.  Figure \ref{fig:samehighqualityexamples} shows visual questions that require \textit{color recognition} tend to share the same grounding to a visual question.  

\section{Dataset Analysis}
\subsection{Incorrect Answer. } 
\begin{figure}[t!]
     \centering
     \includegraphics[width=0.32\textwidth]{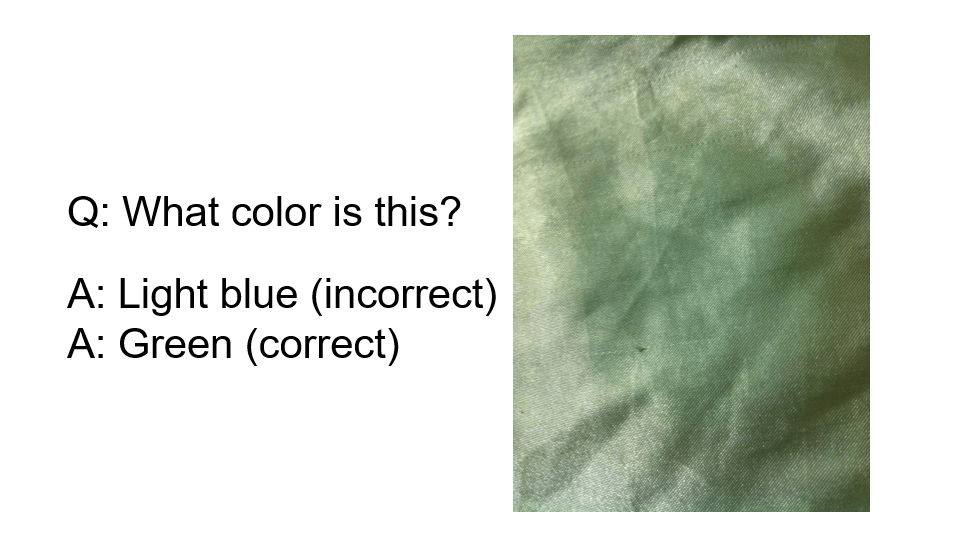}
     \vspace{-1em}
        \caption{An example of a color-related visual question when at least two out of ten people give the same incorrect answer ``light blue". }
    \label{fig:incorrectanswerColor}
\end{figure}

\begin{figure}[t!]
     \centering
     \includegraphics[width=0.48\textwidth]{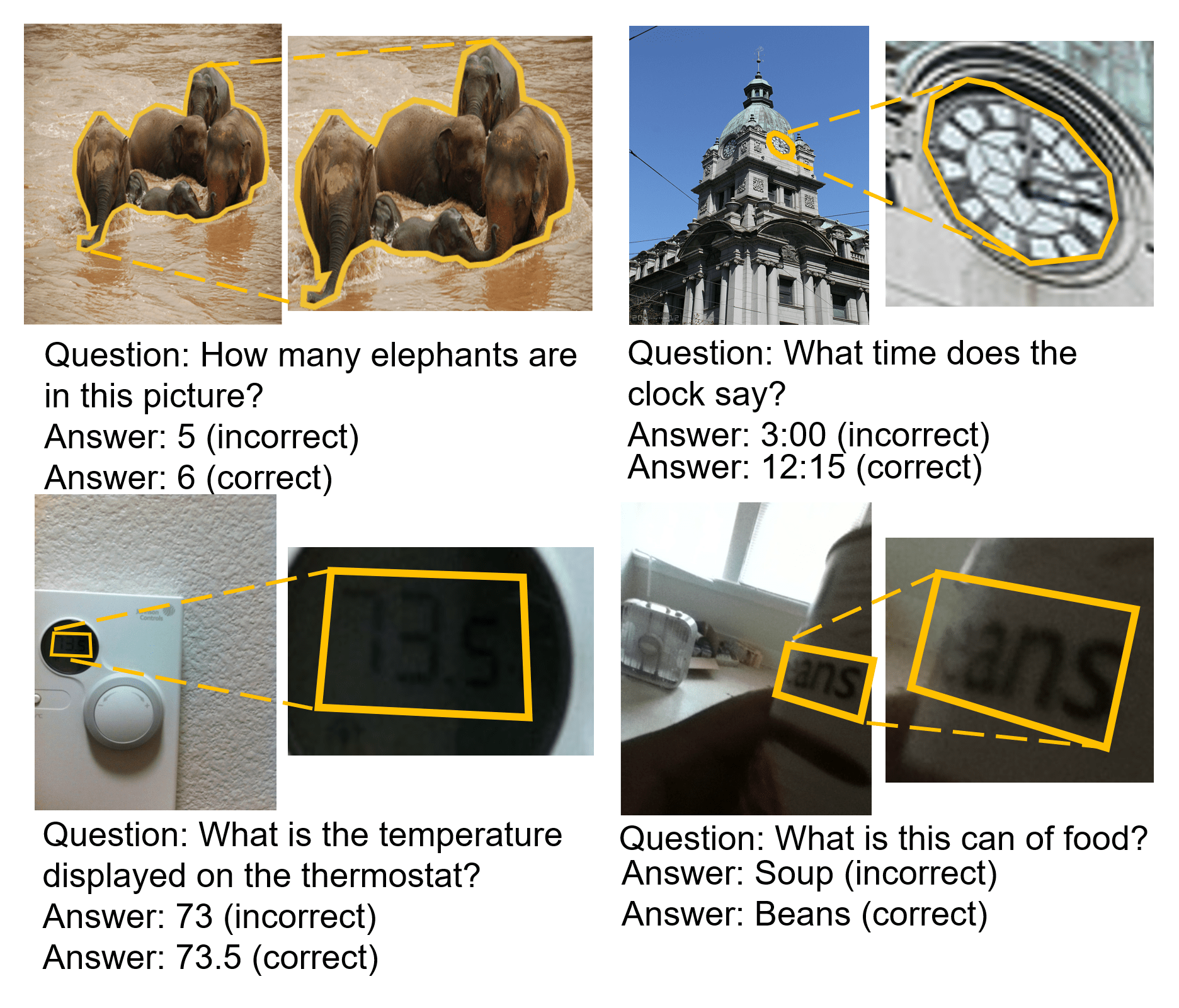}
     \vspace{-2em}
        \caption{Examples show that when regions that lead to the answer are too small to recognize or when the image has low quality, people can answer the visual question incorrectly while achieving agreement (at least two out of ten people give the same incorrect answer). We show the correct answer and the magnified grounding for the correct answer (to the right of the original image) for readers' convenience because, without magnifying, some regions that lead to the answers are too small/too low quality to tell whether the provided answers are correct or not.}
    \label{fig:incorrectanswer}
\end{figure}

Even though we define a valid answer as at least two out of ten people agreeing on that answer, 
we find that 29\% of answers (17,719 out of 60,526) are labeled as incorrect by at least one worker; i.e.,  3,309 out of 20,930 answers from VizWiz-VQA dataset and 14,410 out of 39,596 answers from VQAv2.
From inspection of some of these answers, the reasons why answers are deemed incorrect are (1) regions are too small to recognize, (2) images are too low quality to recognize the content (e.g., too dark or too blurred), and (3) similar colors. For example, an image showing a green cloth while some people say it is light blue is shown in Figure \ref{fig:incorrectanswerColor}).  Examples of incorrect answers are also shown in Figure \ref{fig:incorrectanswer}. Since it is hard to recognize if an answer is correct or not with the low-quality images or small groundings (e.g., the clock region is too small to tell if it is 3:30 or 12:15), we also show the correct answer and its magnified grounding for readers' convenience.

To facilitate future work, we will share the metadata indicating which answers are ``incorrect" as part of publicly-releasing our VQA-AnswerTherapy dataset. Potential use cases for identifying incorrect answers include (1) verifying provided answers in the existing VQA datasets [2, 13], which can lead to cleaner VQA datasets and (2) indicating when the model might perform even better than humans: it might be easier for the model to recognize small regions without magnifying regions and the model can also lighten, darken, or deblur images when needed. Given that a large percentage of flagged incorrect answers exist in both the VizWiz-VQA dataset [13] and the VQAv2 dataset [2], we encourage future work to explore this topic more. 

\subsection{No Polygon and Multiple Polygons. }
\begin{figure}[tph]
     \centering
     \includegraphics[width=0.47\textwidth]{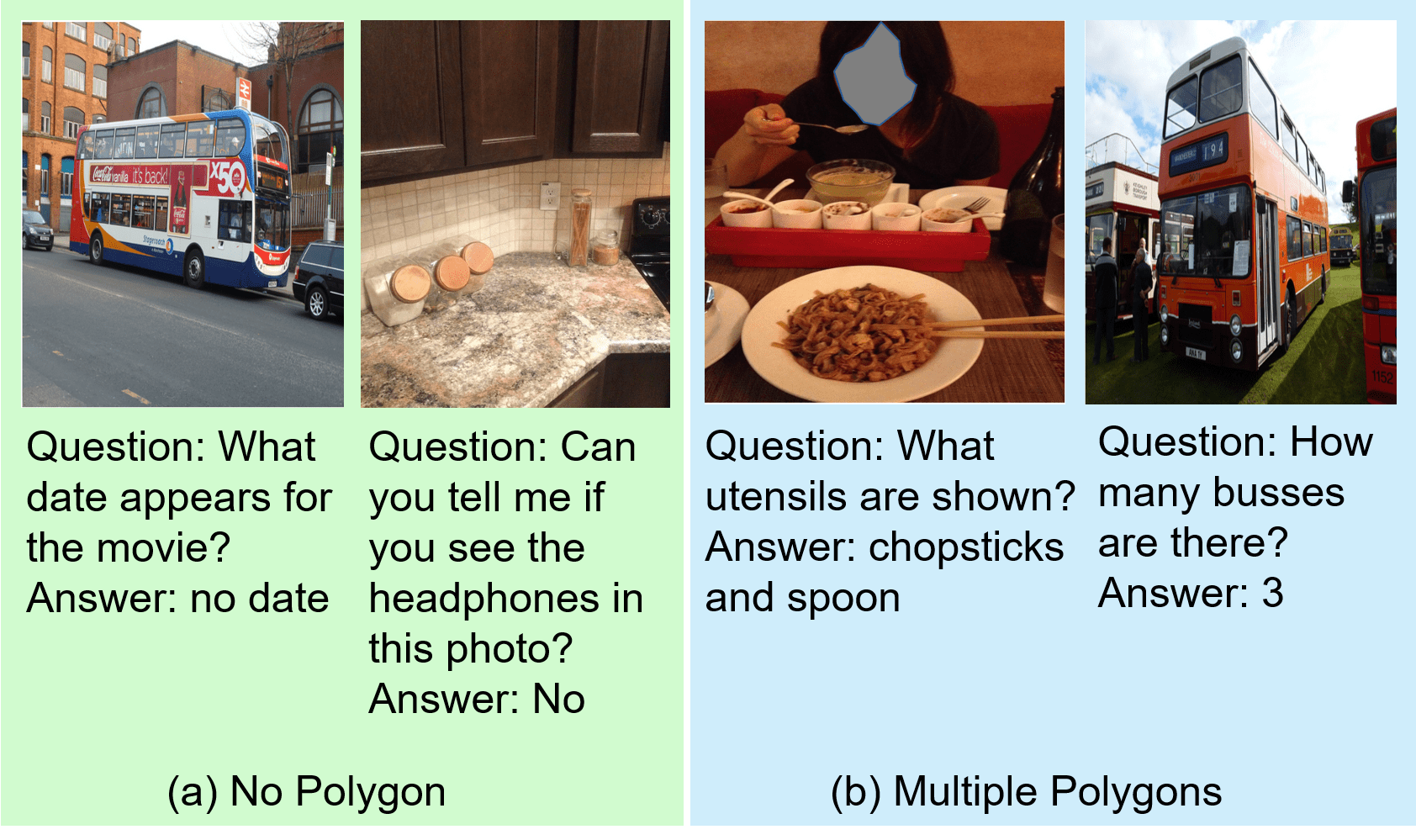}
     \vspace{-1em}
        \caption{Visual questions that (a) have no answer grounding (i.e, need no polygons) and (b) need more than one polygon for the answer grounding.  }
    \label{fig:nomulti}
\end{figure}

Recall that when we collect the data, in step 2 we asked workers to indicate ``how many polygons are needed to locate the region that the answer is referring to". We show some visual questions when people select ``no polygon is needed" and ``multiple polygons are needed" in Figure \ref{fig:nomulti}. 

\subsection{Grounding Agreement. }
Recall that two answer grounding annotations were collected for each unique answer per visual question from two crowdworkers. 
\begin{figure}[tph]
         \centering
     \includegraphics[width=0.5\textwidth]{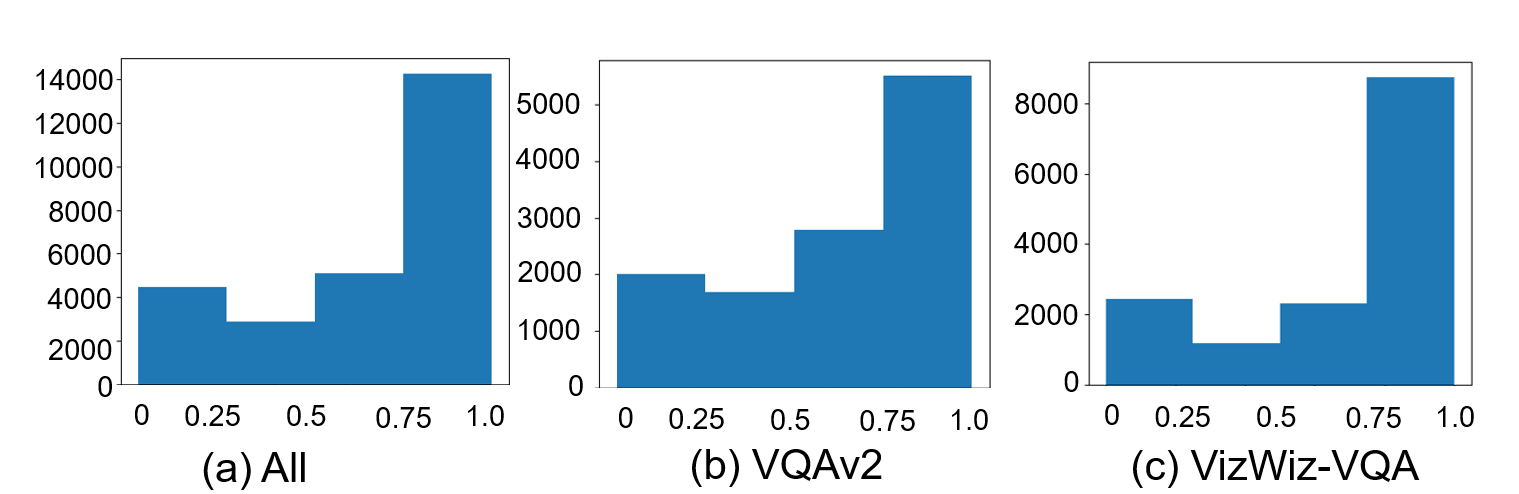}
     \vspace{-2em}
        \caption{ Histogram of IoU scores indicating similarity between each pair of answer groundings per visual question. The majority have a high agreement, in the range between 0.8 and 1.0. }
        \label{fig:alignment}
\end{figure}

\begin{figure}[tph]
     \centering
     \includegraphics[width=0.5\textwidth]{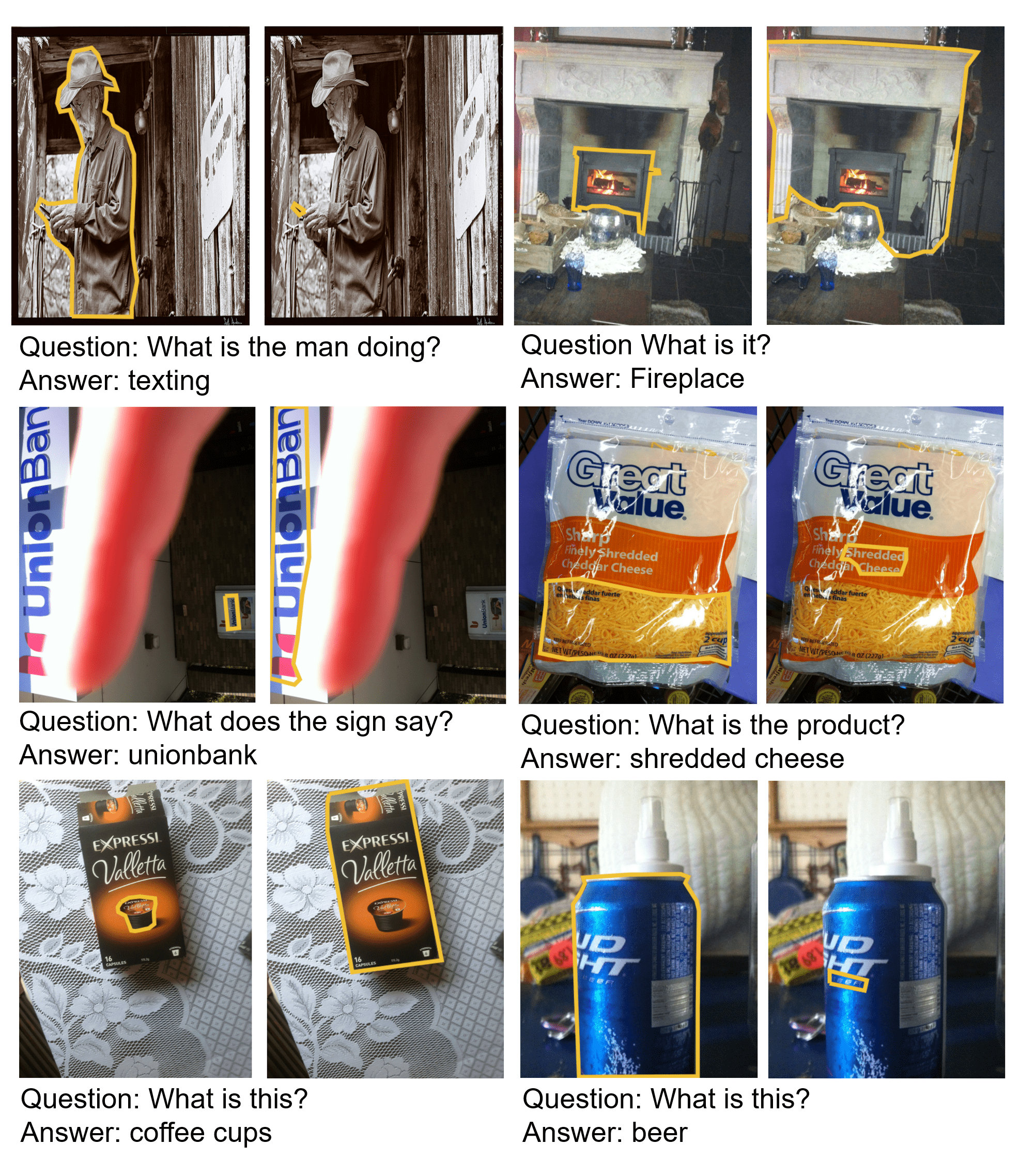}
     \vspace{-2.5em}
        \caption{Examples of low alignment between two workers' annotations because of ambiguous or redundant information where different regions can independently indicate the same answer.  }
    \label{fig:groundinglowalignment}
\end{figure}

We show a histogram of grounding alignment between two crowdworkers across the 26,682 unique image-question-answer triples in Figure \ref{fig:alignment}.
The majority (53\%, 14,262 out of 26,682) of the IoU scores are between 0.75 and 1.0, $\sim$20\% (5,101 out of 26,682) between 0.75 and 0.5, $\sim$10\%(2,865 out of 26,682) between 0.5 and 0.25, and 17\% (4,453 out of 26,682) lie between 0.25 and 0. We attribute grounding misalignments largely to the grounding being ambiguous, as exemplified in the first row of Figure \ref{fig:groundinglowalignment}, and redundant information in the image where different regions can independently indicate the same answer, as exemplified in the second row of Figure \ref{fig:groundinglowalignment}.

The grounding differences from different workers highlighted a few questions that we leave for future work: (1) When grounding an answer, should we ground all the information (both the explicit information and the implicit information) that leads to the answer, or just explicit information?, (2)
Should we ground all the information or just part of information (e.g., many regions independently lead to the same answer and we just ground the most obvious one) if part of the information is already sufficient?, (3) When workers draw regions that are highly aligned with each other, which grounding should we select? 

\subsection{Reconciling Redundant Annotations.}
 As mentioned in the main paper, during the annotation process, we allow workers to select each answer if the answer has been located in one of the previously drawn regions (See Figure \ref{fig:instruction} Step 3 - Option (a) and Figure \ref{fig:userinterface} Step 3). Then we selected the larger grounding from two groundings if the two groundings' alignment is larger than 0.75. We observe that frequently  (93\%, i.e., for 5,459 out of 5,825 visual questions), the selected answer grounding for different answers to one visual question are from the same worker (recall though that the annotations across different visual questions still can come from different workers). For VizWiz-VQA, 3153 visual questions each have all answer groundings coming from the same worker and 289 from different workers. For VQAv2, 2306 visual questions each have all answer groundings coming from the same worker and 77 from different workers.  These facts highlight that a visual question's different answer groundings can all be identical and so have an IoU = 1.0.  

We decide whether the answer groundings are based on the same regions by calculating IoU scores for every possible answer grounding pair per visual question and checking if all of the grounding answer pairs have an IoU score larger than 0.9. If their overlap is larger than 0.9, we believe this visual question has the same grounding for all answers. We chose an IoU threshold less than 1.0 to accommodate the 7\% of visual questions where different answer groundings for the same visual question came from different workers.  We also report in Table \ref{table:thresholdSameGrounding} the number of visual questions identified as having a single versus multiple groundings when using different IoU thresholds between 0.9 and 1.0. The results show similar outcomes when using different thresholds. 

\begin{table}[]
\begin{tabular}{lllllll}
\hline
& \multicolumn{2}{l}{All} & \multicolumn{2}{l}{VQAv2}                                    & \multicolumn{2}{l}{VizWiz-VQA}         \\ \hline
IoU           & Single       & Mult       & Single                                                & Mult & Single                        & Mult \\ \hline
0.7            & 5027       & 798        &  2245 & 138  &2782 & 660  \\
0.75           & 4992       & 833        & 2243                                                & 140  & 2749                        & 693  \\
0.8           & 4957       & 868        & 2238                                                & 145  & 2719                        & 723  \\
0.85           & 4932       & 893        & 2235                                                & 148  & 2697                        & 745  \\
0.9            & 4909       & 916        & 2228                                                & 155  & 2681                        & 761  \\
0.95           & 4896       & 929        & 2225                                                & 158  & 2671                        & 771  \\
1             & 4889       & 936        & 2223                                                & 160  & 2666                        & 776  \\ \hline
\end{tabular}
\caption{ Number of VQAs with a single grounding and multiple (Mult) groundings under different IoU thresholds. }
\label{table:thresholdSameGrounding}
\end{table}

\subsection{Four Grounding Relationships. }
\begin{figure}[tph]
     \centering
     \includegraphics[width=0.45\textwidth]{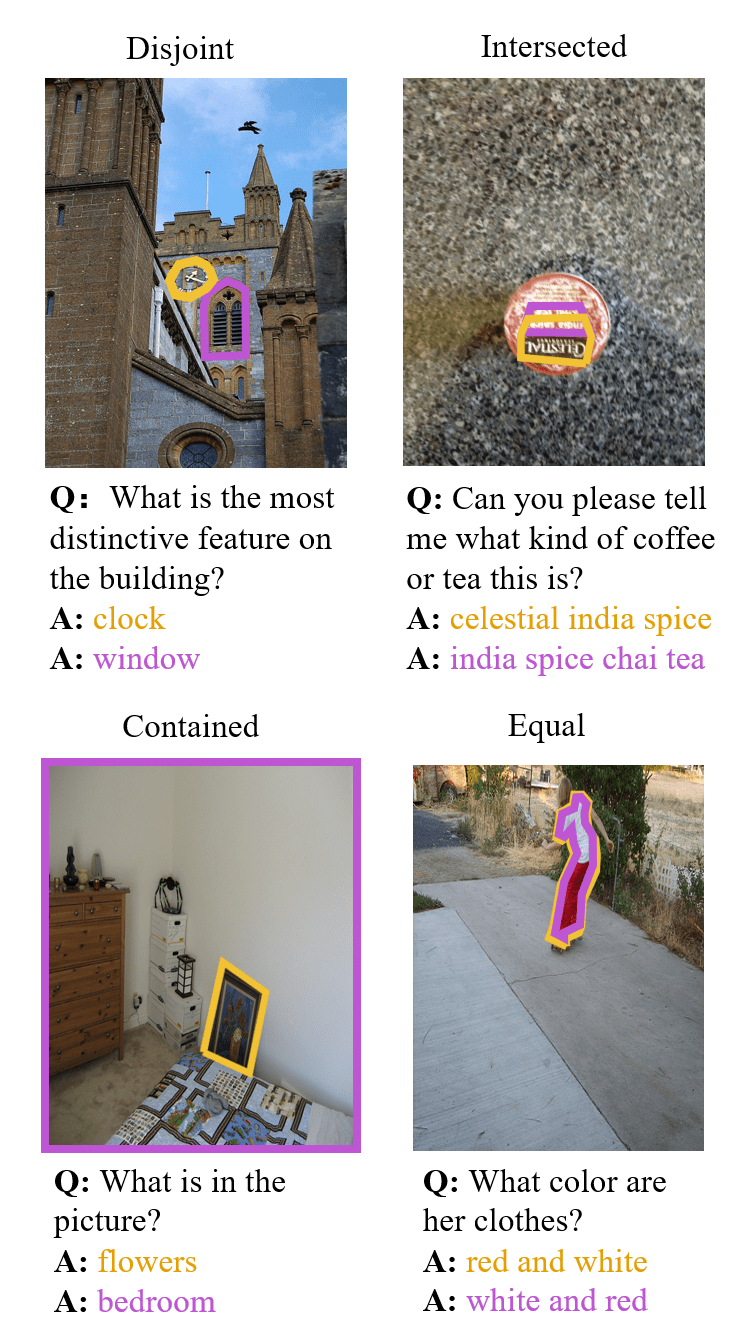}
     \vspace{-1em}
        \caption{For each visual
question, we flag which relationship types arise between every possible answer grounding pair from the following options: disjoint, equal, contained, and intersected.  }
    \label{fig:4relationship}. 
\end{figure}

We visualize four kinds of relationships, i.e., disjoint, equal, contained, and intersected, between every possible answer grounding pair in Figure \ref{fig:4relationship}. These exemplify that visual questions needing \textit{object recognition} tend to have disjoint or contained relationships, visual questions needing \textit{text recognition} tend to have intersected relationships, and visual questions needing \textit{color recognition} tend to have an equal relationship.  

\subsection{Most Common Answers.}
\begin{figure}[tph]
     \centering
     \includegraphics[width=0.47\textwidth]{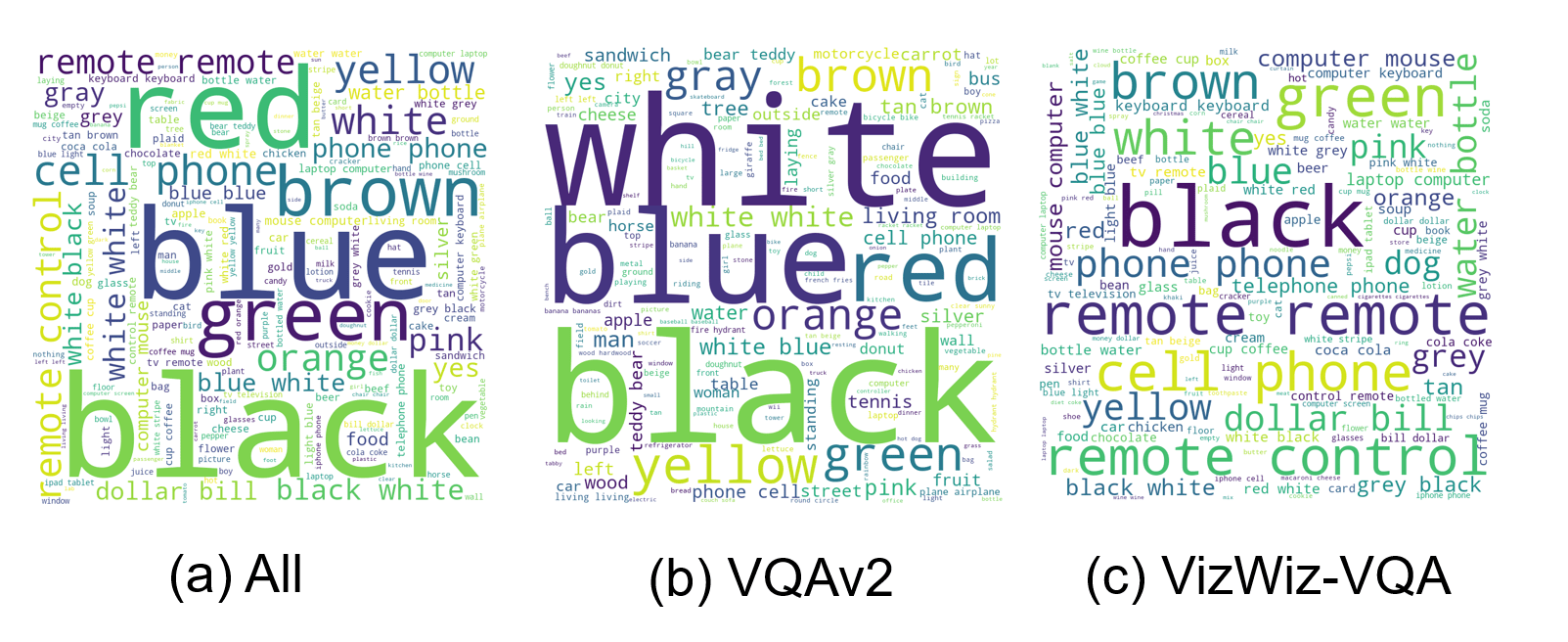}
     \vspace{-1em}
        \caption{Most common answers for visual questions that have the same groundings for all unique answers.  }
    \label{fig:answer}
\end{figure}

Due to space constraints, we provide the analysis of the most common answers that co-occur with a single grounding here. We obtain the most common answers following a similar process as used to obtain the most common questions in the main paper.  The top five common answers for the VQA-AnswerTherapy dataset that co-occur with a single grounding are `white', `phone', `blue', `black', and `brown'. The top five common answers for VQAv2 are `white', `brown', `black', `gray', and `blue'. The top five common answers for VizWiz-VQA are `phone', `grey',  `blue',  `remote', and `remote control'.  We show the WordCloud for the common answers that lead to the same answer grounding for VQA-AnswerTherapy as well as for the VizWiz-VQA and VQAv2 datasets independently in Figure \ref{fig:answer}.  These findings reinforce our conclusion in the main paper that visual questions requiring object or color recognition skills tend to share the same groundings.

\section{Algorithm Benchmarking}

\paragraph{Experimental Details for Single Answer Grounding Challenge. }
We used an AdamW optimizer with a learning rate of 0.00005 and fine-tuned ViLT on the VizWiz-VQA and VQAv2 datasets for 20 epochs.  

For mPLUG-Owl, we did preliminary testing with four different prompts and selected the best one: 

            ```The following is a conversation between a curious human and AI assistant. The assistant only replies ``YES" or ``NO" to the user's questions.\\
            Human: \textless image\textgreater \\
            Human: What are all plausible answers to the question \textless INSERT QUESTION VARIABLE\textgreater?\\
            Human: Do all plausible answers to the questions  \textless INSERT QUESTION VARIABLE\textgreater  indicate the same visual content in this image? Reply ``YES" or ``NO".\\
            AI: '''. 

The responses from mPLUG-Owl were typically either ``yes" or ``no" followed by a reason, (even though the model was instructed not to respond with reason). We converted the first three characters of each response to lowercase and then compared them to the ground truth to see if there is a match. If the response is anything other than ``yes" or ``no," we disregard it as it cannot be reflected in precision or recall.  There are 10 out of 496 samples that don't have "yes" or "no" as their first three characters in the VQAv2 dataset, and there are 16 such instances out of 889 samples in the VizWiz-VQA dataset.

\paragraph{Experimental Details for Answer(s) Grounding Challenge. }
For SeqTR model, we used the pre-trained RefCOCOg weights from the SeqTR author's repository (https://github.com/sean-zhuh/SeqTR) and fine-tuned it for 5 epochs following the author's guidelines.
For the UNINEXT model, we used UNINEXT’s second stage pre-trained weights, which were top-performing for COCO detection and segmentation (verified by author). Of note, UNINEXT is also pretrained on RefCOCO and so was exposed to the COCO images utilized in our dataset. For SEEM model, we used the SEEM-FOCAL-V1 checkpoint from author's repository (https://github.com/UX-Decoder/Segment-Everything-Everywhere-All-At-Once). SEEM is also pre-trained on RefCOCO and COCO2017 and so was also exposed to the COCO images utilized in our dataset. 

\subsection{Performance of Three Models for Answer(s) Grounding Challenge with IoU-PQ Metric}
 We show the IoU-PQ performance in Table \ref{table:miouPQ}, the results and observations are highly aligned with the mIoU metric reported in our main paper. 
\begin{table}[t!]
\centering
\begin{tabular}{cccc}
\hline
\textbf{Models}  &\textbf{All} & \textbf{VQAv2} & \textbf{VizWiz-VQA}\\\hline

SeqTR (I+Q+A)   & \textbf{66.26}  &  \textbf{64.34}  & \textbf{67.33}\\
SeqTR (I+Q) &  61.62&   58.30&  63.47\\
SeqTR (I+A) &  62.91& 57.97   &  65.67\\\midrule
SEEM (I+Q+A)    & \textbf{53.15} & \textbf{50.13}   & \textbf{54.84} \\
SEEM (I+Q)         &  44.65&  44.39   & 44.80 \\
SEEM (I+A)          & 51.64&  46.50 &  54.51 \\\midrule
UNINEXT (I+Q+A) &\textbf{48.39}  & \textbf{42.28} &  \textbf{59.34}\\
UNINEXT (I+Q)  & 45.88 &  40.76 & 55.06  \\
UNINEXT (I+A) & 47.45  &  41.26  &58.55 \\

\hline

\end{tabular}
\vspace{-0.25em}\caption{mIoU-PQ Performance of three models on our dataset.}
\label{table:miouPQ}
\end{table}
\subsection{Answer(s) Grounding: Qualitative Results for Model Benchmarking.}
\begin{figure*}[h!]
     \centering
     \includegraphics[width=1\textwidth]{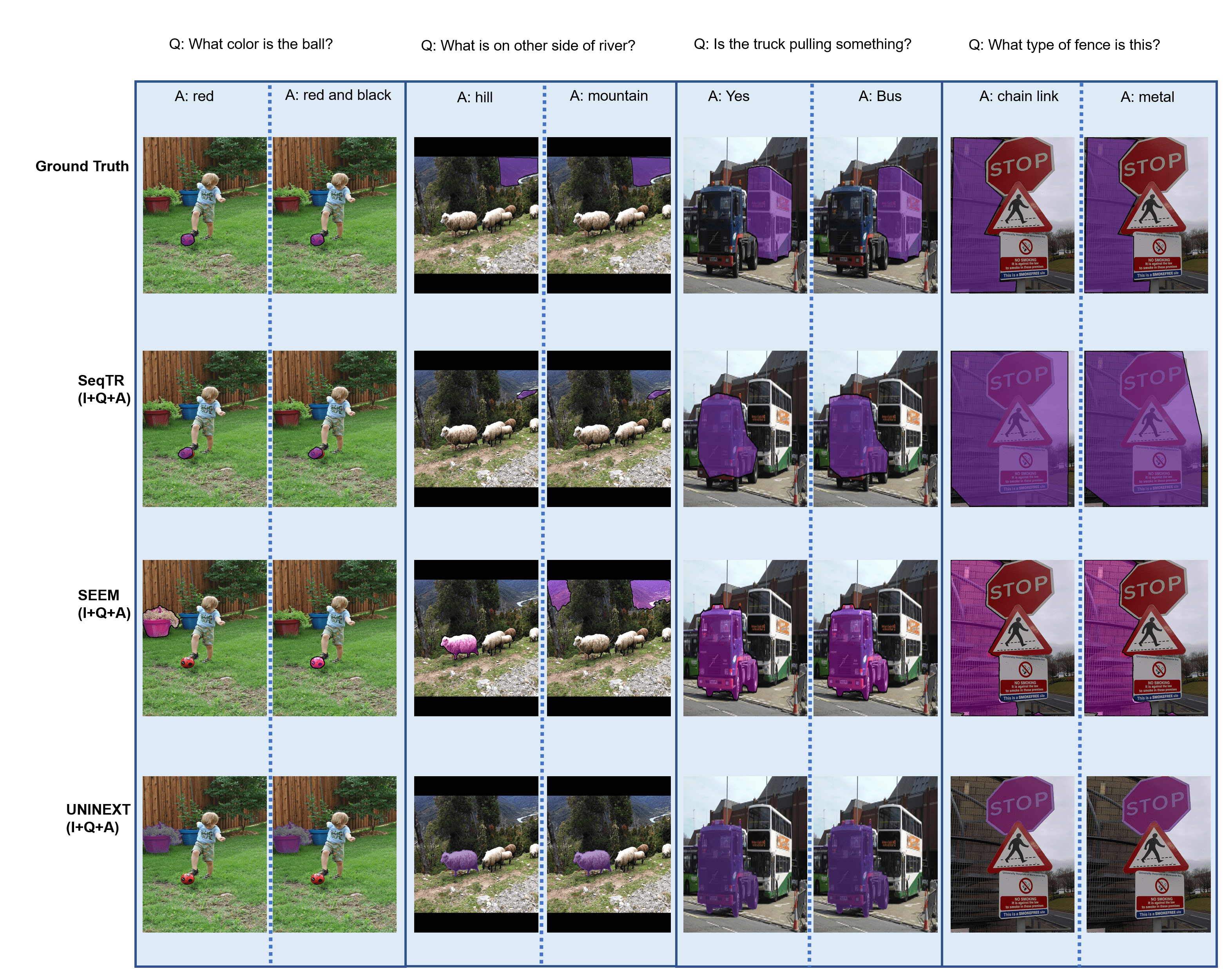}
     \vspace{-1em}
        \caption{Qualitative results for models tested on visual questions with different answers leading to \textbf{same groundings}. Image sources are VQAv2 datasets (in the blue background). For each visual question, the  first row shows the ground truth grounding area, the second, third, and fourth row show groundings generated by different models. Each column shows the grounding for an answer.  }
    \label{fig:algorithmSameVQA}
\end{figure*}

\begin{figure*}[h!]
     \centering
     \includegraphics[width=1\textwidth]{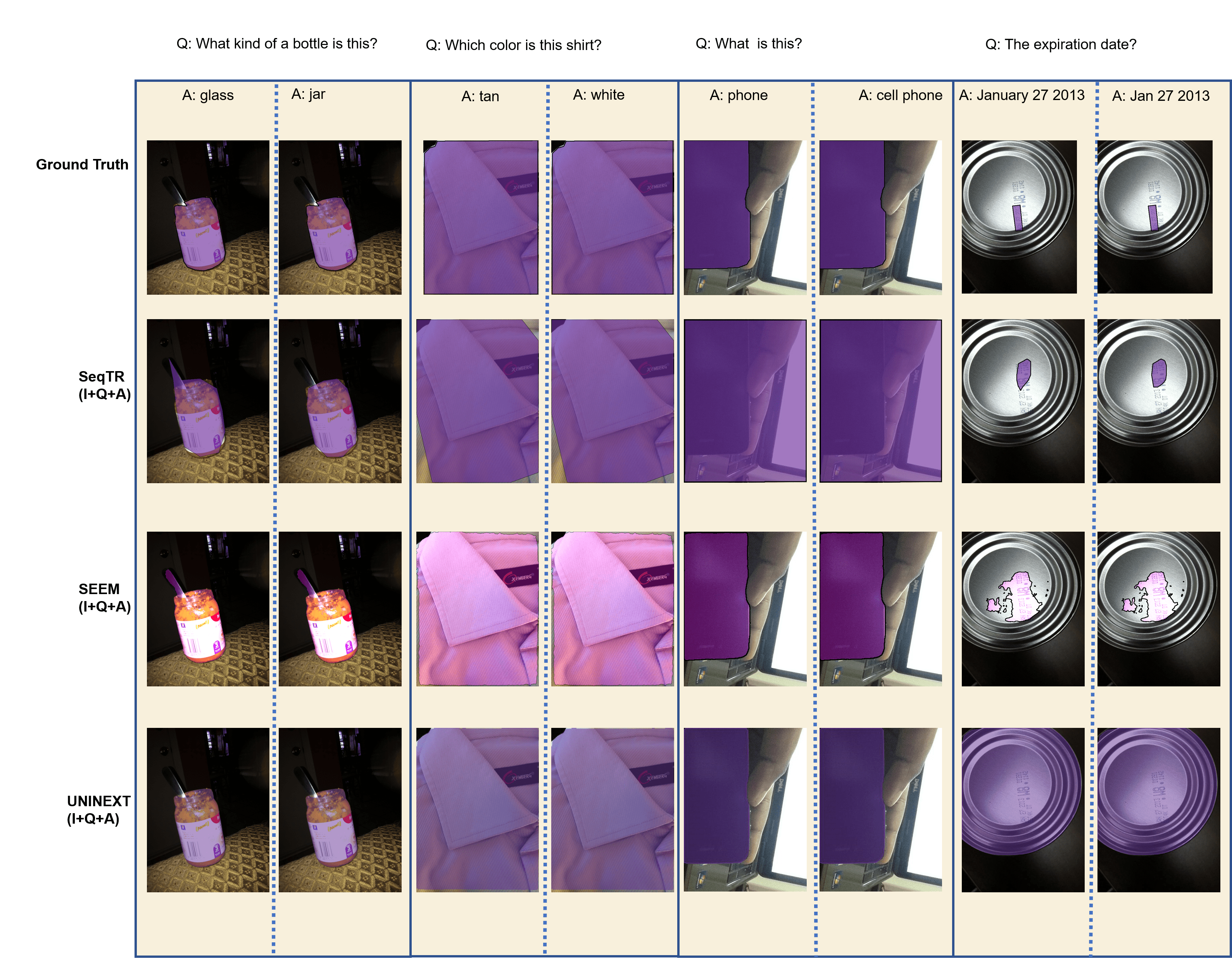}
     \vspace{-1em}
        \caption{Qualitative results for models tested on visual questions with different answers leading to \textbf{same groundings}. Image sources are VizWiz-VQA datasets (in the yellow background). For each visual question, the  first row shows the ground truth grounding area, the second, third, and fourth row show groundings generated by different models. Each column shows the grounding for an answer. }
    \label{fig:algorithmSameVizWiz}
\end{figure*}

\begin{figure*}[h!]
     \centering
     \includegraphics[width=1\textwidth]{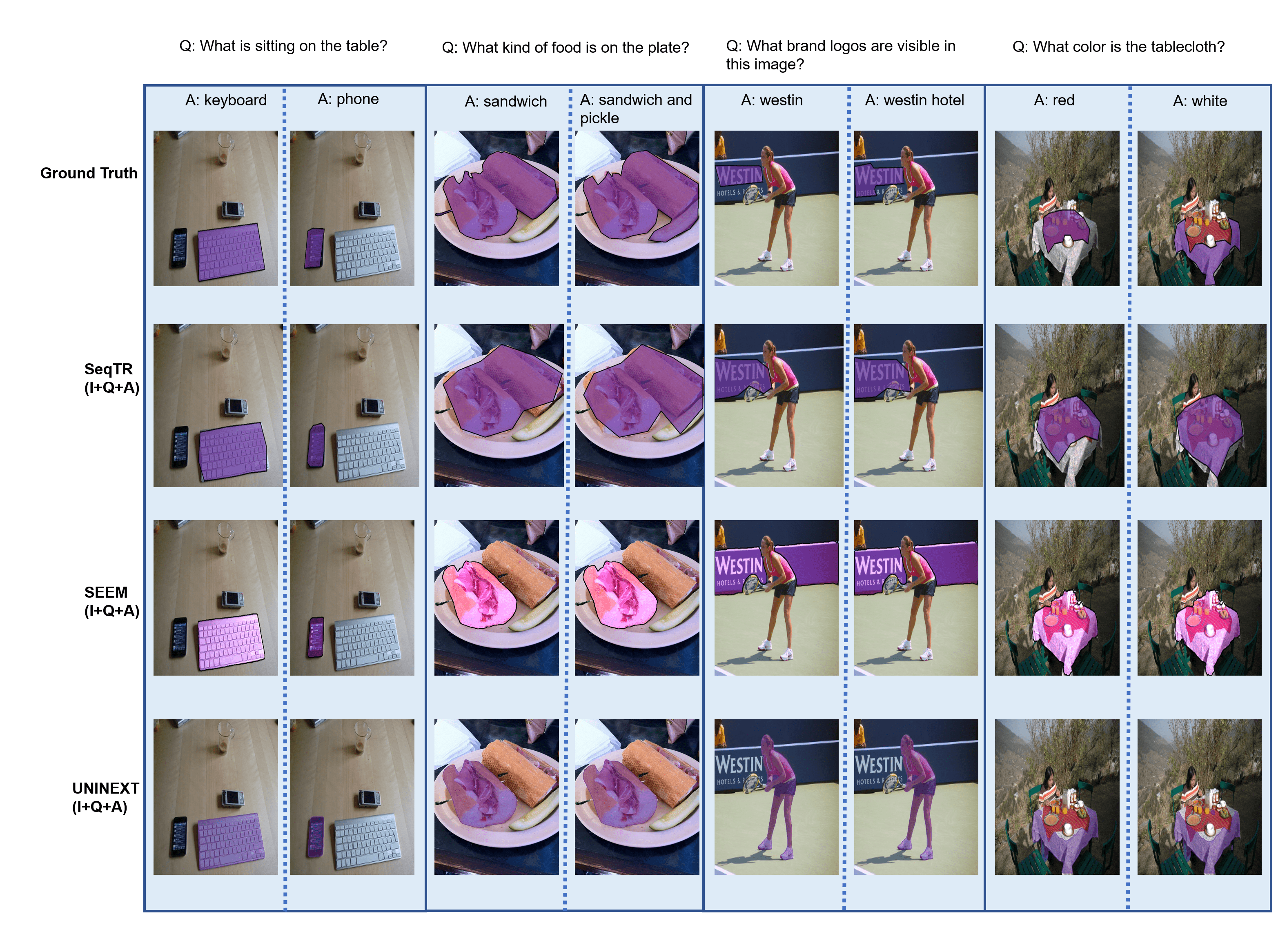}
     \vspace{-1em}
        \caption{Qualitative results for models tested on visual questions with different answers leading to \textbf{different groundings}. Image sources are VQAv2 datasets (in the blue background). For each visual question, the first row shows the ground truth grounding area, and the rest of the rows show the models' predicted area. Each column shows the grounding for an answer. }
    \label{fig:diffgroundingvqa}
\end{figure*}

\begin{figure*}[h!]
     \centering
     \includegraphics[width=1\textwidth]{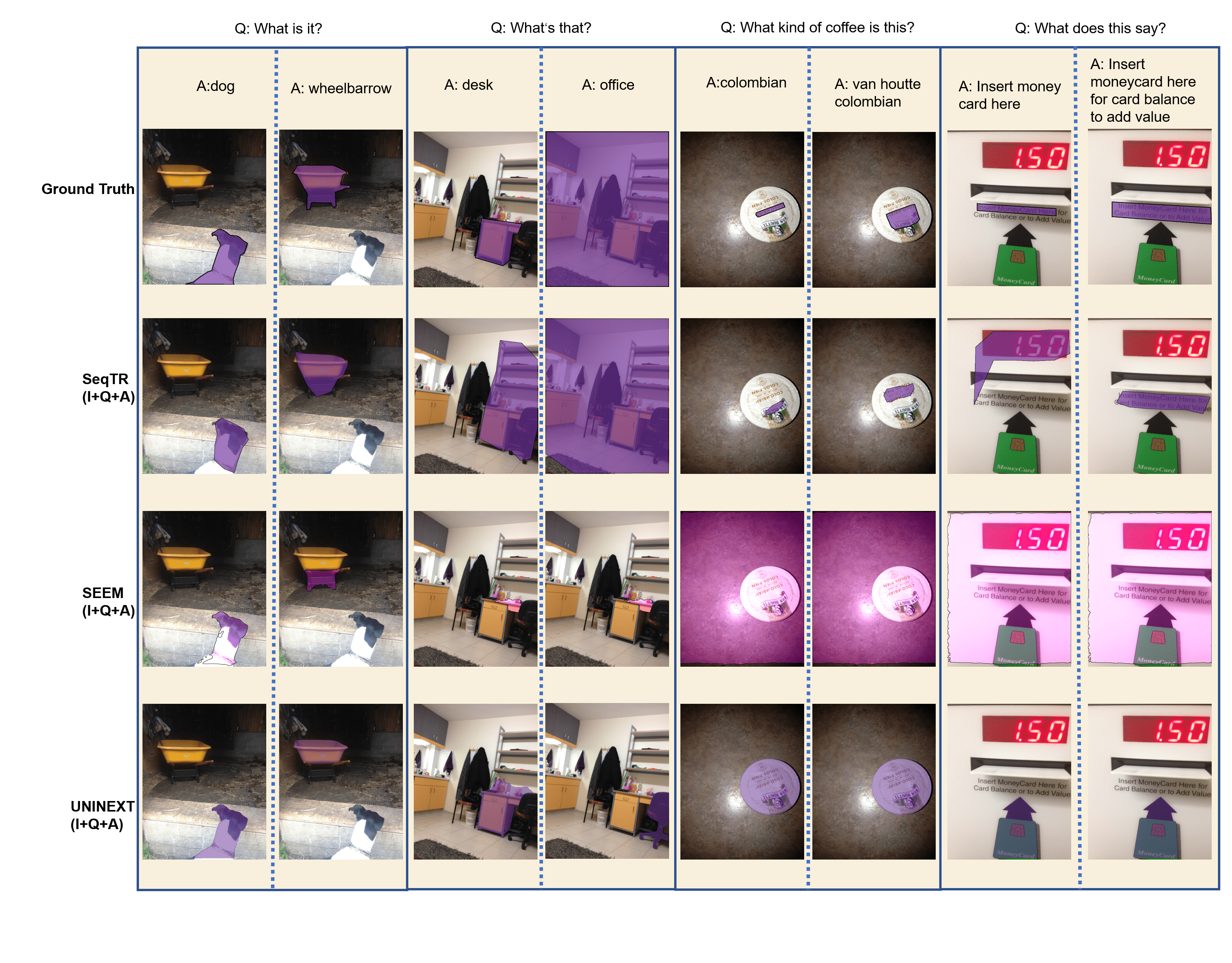}
     \vspace{-1em}
        \caption{Qualitative results for models tested on visual questions with different answers leading to \textbf{multiple groundings}. Image sources are VizWiz-VQA (in the yellow background). For each visual question, the first row shows the ground truth grounding area, and the rest of the rows show the models' predicted area. Each column shows the grounding for an answer.  }
    \label{fig:diffgroundingVizwiz}
\end{figure*}

We provide additional qualitative results here for the Answer(s) Grounding task for the top-performing set-up where we feed models the image, question, and answer. . Examples are provided in Figures \ref{fig:algorithmSameVQA}, \ref{fig:algorithmSameVizWiz}, \ref{fig:diffgroundingvqa}, and \ref{fig:diffgroundingVizwiz}.  

Figures \ref{fig:algorithmSameVQA} and \ref{fig:algorithmSameVizWiz} show visual questions with different answers that lead to the \textbf{same groundings}. Overall, we observe that models can predict well for this case, particularly when grounding a single dominant object on a relatively simple background. However, if the picture is captured from an unusual perspective or shows multiple objects (e.g., \ref{fig:algorithmSameVQA} ``Is the truck pulling something"), models can fail. We also observe that though the answers are referring to the same region, the model's predictions for different answers sometimes can differ. This is exemplified in Figure \ref{fig:algorithmSameVQA}'s column 1 for SEEM(I+Q+A) (``What color is the ball") and column 2 for SEEM (I+Q+A) (``What is on other side of river").  

Figures \ref{fig:diffgroundingVizwiz} and \ref{fig:diffgroundingvqa} show the qualitative results for the models tested on visual questions with different answers that lead to \textbf{multiple groundings}. Though different answers can refer to different regions, the model's predictions for different answers are sometimes the same. The model might perform better when identifying common objects when the camera directly faces the object (e.g., shown in Figure \ref{fig:diffgroundingvqa}'s column 1 (``What is sitting on the table?")  and worse when the content of interest is captured from other perspectives (e.g.,  Figure \ref{fig:diffgroundingVizwiz}'s column 2 (``What's that?"). The model also fails to distinguish regions for different text related answers, as exemplified in Figure \ref{fig:diffgroundingvqa}'s column 3 (``What brand logos are visible in this image?") and Figure \ref{fig:diffgroundingVizwiz}'s column 3 (``What kind of coffee is this?") and column 4 (``What does this say?").


\end{document}